\documentclass[10pt,twocolumn,letterpaper]{article}

\usepackage[pagenumbers]{iccv} % To force page numbers, e.g. for an arXiv version

\usepackage{array}
\usepackage{booktabs}
\usepackage{multirow}
\usepackage{xcolor}

\definecolor{iccvblue}{rgb}{0.21,0.49,0.74}
\usepackage[pagebackref,breaklinks,colorlinks,allcolors=iccvblue]{hyperref}

\def\paperID{*} % *** Enter the Paper ID here
\def\confName{ICCV}
\def\confYear{2025}

\title{Hybrid Agents for Image Restoration}

\author{Bingchen Li \quad Xin Li\thanks{Corresponding Author} \quad Yiting Lu \quad Zhibo Chen\\
University of Science and Technology of China \\
{\tt\small \{lbc31415926, luyt31415\}@mail.ustc.edu.cn, \{xin.li, chenzhibo\}@ustc.edu.cn}
}

\begin{document}
\maketitle
\begin{abstract}
Existing Image Restoration (IR) studies typically focus on task-specific or universal modes individually, relying on the mode selection of users and lacking the cooperation between multiple task-specific/universal restoration modes. This leads to insufficient interaction for unprofessional users and limits their restoration capability for complicated real-world applications. In this work, we present HybridAgent, intending to incorporate multiple restoration modes into a unified image restoration model and achieve intelligent and efficient user interaction through our proposed hybrid agents. Concretely, we propose the hybrid rule of fast, slow, and feedback restoration agents. Here, the slow restoration agent optimizes the powerful multimodal large language model (MLLM) with our proposed instruction-tuning dataset to identify degradations within images with ambiguous user prompts and invokes proper restoration tools accordingly. The fast restoration agent is designed based on a lightweight large language model (LLM) via in-context learning to understand the user prompts with simple and clear requirements, which can obviate the unnecessary time/resource costs of MLLM. 
Moreover, we introduce the mixed distortion removal mode for our HybridAgents, which is crucial but not concerned in previous agent-based works. It can effectively prevent the error propagation of step-by-step image restoration and largely improve the efficiency of the agent system. We validate the effectiveness of HybridAgent with both synthetic and real-world IR tasks.

\end{abstract}    
\section{Introduction}
\label{sec:intro}
Image restoration (IR) has long been a popular topic in the low-level research fields~\cite{liang2021swinir,HAT,zamir2022restormer,xia2023diffir,li2023efficientIR}, which aims to restore low-quality inputs (LQs) into high-quality outputs (HQs). Early IR works primarily focus on model design for diverse simple single restoration tasks, such as image denoising~\cite{denoise1,denoise2,denoise3,denoise4,denoise5}, deblurring~\cite{deblur1,deblur2,deblur3,deblur4,deblur5},  compression artifacts removal~\cite{dejpeg1,dejpeg2,dejpeg3,dejpeg4,arcnn}, and super-resolution~\cite{bsrgan,HAT,realesrgan,srcnn,srgan}. 
However, deploying multiple IR models simultaneously to handle diverse degradations in real-world applications incurs significant resource costs. Recently, all-in-one/universal image restoration models (AIR)~\cite{airnet,potlapalli2023promptir,duan2025uniprocessor,conde2024instructIR} have emerged to address diverse image restoration tasks within a single IR network by learning degradation descriptors through prompt learning or instruction tuning.

\begin{figure}[t]
    \centering
    \includegraphics[width=1\linewidth]{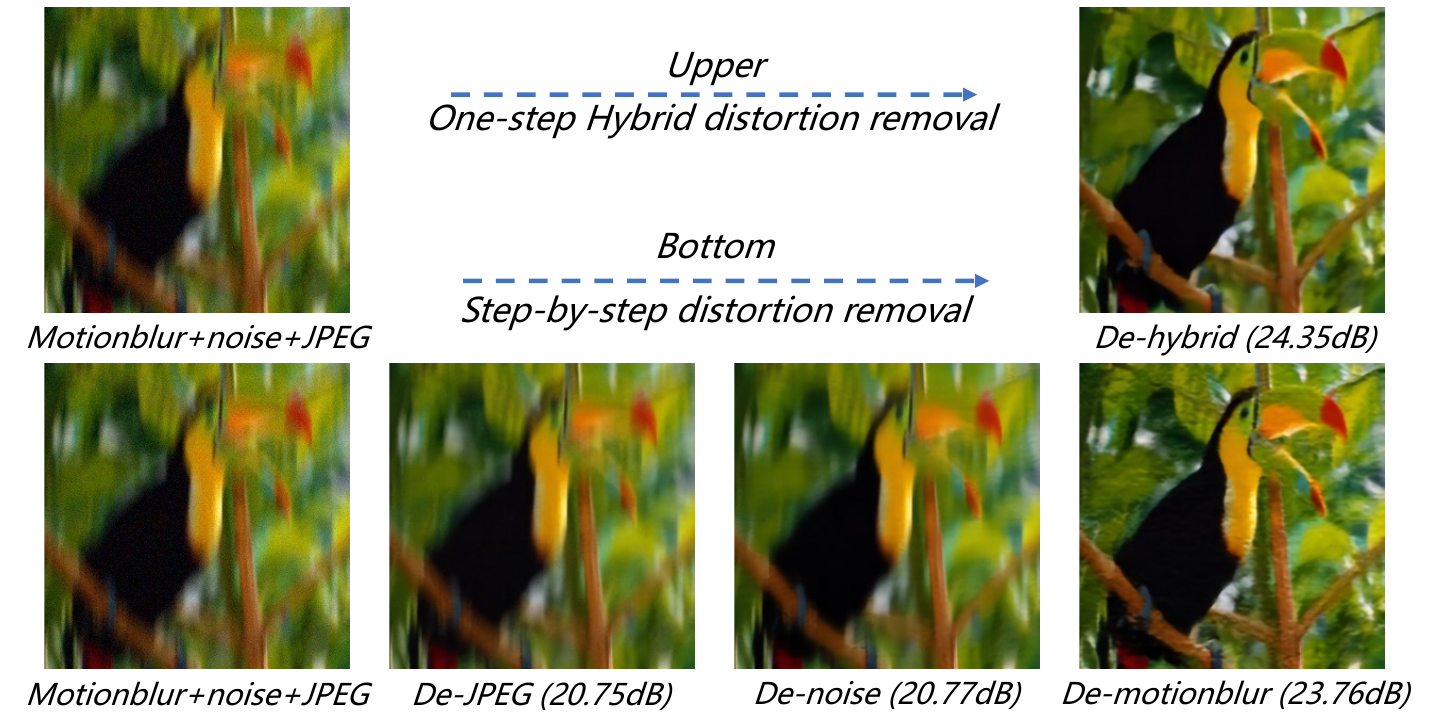}
    \caption{Removing hybrid distortions step-by-step will cause suboptimal results due to distortion entanglement.}
    \label{fig:intro}
\end{figure}

Despite the abundant restoration modes in existing IR studies, the restoration process still depends on professional users for mode selection when dealing with complex real-world degradations, lacking flexible cooperation between different restoration modes. For instance, an untrained user who does not understand terms like ``noise'' or ``blur'' may struggle to select the appropriate restoration tools in a step-by-step manner. To eliminate this, language-based interaction and planning strategies have been proposed to support automatic image restoration~\cite{chen2024restoreagent,agenticIR} with multimodal large language models (MLLMs). However, existing methods suffer from two key challenges: (i) a single interactive agent lacks adaptability to varying task complexities, as even simple and clear instructions, such as ``please remove the noise in the image,'' are still processed using heavyweight MLLMs, leading to unnecessary computational and time costs; (ii) error propagation occurs when step-by-step single degradation removal tools are applied to real-world hybrid degradations, as illustrated in Fig.~\ref{fig:intro}.

In this work, we propose HybridAgent, a novel interactive paradigm for universal image restoration that addresses the aforementioned challenges through two key innovations: (i) the hybrid restoration agents, consisting of fast, slow and feedback agents to balance efficiency and effectiveness during automatic restoration process; and (ii) mixed distortion removal tools integrated with single distortion removal tools to enhance adaptability and prevent error propagation during restoration process. 

Specifically, we apply an MLLM as our slow restoration agent (\textbf{SlowAgent}) due to its powerful understanding, reasoning, and decision-making capabilities. To adapt SlowAgent to IR, we construct an instruction-tuning dataset designed for complex image restoration tasks. We then fine-tune the SlowAgent, equipping it with strong task understanding, distortion identification, and restoration mode/tool planning capabilities. However, user interaction requirements are not always complex, and not every task necessitates the agent to identify degradations within images. For simple interactive IR tasks (\eg, ``please remove noise''), applying an MLLM would be inefficient and resource-intensive. To enable flexible and efficient interaction for different IR requirements, we introduce the fast restoration agent (\textbf{FastAgent}) by leveraging a lightweight large language model (LLM) to determine whether the user prompt is clear and simple. If so, the FastAgent will directly execute the planning for the simple IR task and invoke the corresponding restoration tool. Otherwise, the task is handed over to SlowAgent, which automatically identifies degradations and performs the restoration. Notably, to achieve an automated restoration process, SlowAgent requires external feedback to determine whether the current image still needs further restoration. To better facilitate SlowAgent in completing the restoration process, we further develop a \textbf{FeedbackAgent} to assess whether a restored image is free from degradations. The three restoration agents collaborate to enable the automatic restoration of distorted images, formulating our HybridAgent.

Another essential component of HybridAgent is the restoration tools, which are invoked by the agents during the restoration process. A straightforward approach would be to directly utilize state-of-the-art single-task restoration models as tools~\cite{rl-restore,chen2024restoreagent,agenticIR}. However, this approach fails to exploit shared knowledge across tasks. Moreover, distributional discrepancies between different models may cause error propagation during the restoration process. To address this, we propose a three-stage training paradigm for constructing our restoration tools, which not only enables the tools to effectively preserve common knowledge shared across different restoration tasks but also provides an intuitive approach for building both single distortion and mixed distortion removal tools. In the first stage, following the pretraining-finetuning paradigm, we train a foundation restoration model via a multi-task learning scheme~\cite{airnet,potlapalli2023promptir} to learn the task-shared knowledge. Then, in the second stage, we leverage Low-Rank Adaptation (LoRA)~\cite{hu2021lora} to efficiently fine-tune single distortion removal tools from the pre-trained foundation model. Consequently, in the third stage, we further tailor a mixed distortion removal tool based on the pre-trained foundation model with a new set of LoRA weights. Notably, in our three-stage training paradigm, we adopt prompt components~\cite{potlapalli2023promptir,PIP} to implicitly encode distortion information, while employing LoRA to efficiently adapt the pre-trained model to domain shifts caused by different distortions.

Compared to directly adopting existing restoration models as tools, our proposed three-stage training strategy offers a more flexible and efficient way to construct restoration tools. Specifically, when users encounter new restoration tasks, they no longer need to search for specialized restoration models (which may not even exist, such as models specifically designed for VVC compression artifact removal). Instead, users can simply leverage LoRA to efficiently fine-tune the first-stage foundation model for the new task.

To cover the majority of application scenarios, we train all restoration tools and build the instruction dataset based on 10 degradations with various levels, including: noise~\cite{denoise1}, gaussian blur, motion blur~\cite{deblur2}, JPEG~\cite{dejpeg1}, HEVC~\cite{HM}, VVC~\cite{VTM}, rainstreak~\cite{rain100H}, raindrop~\cite{raindrop}, haze~\cite{engin2018cycledehaze}, low light~\cite{wang2018gladnetlow-light}. We incorporate two less-explored compression codecs, HEVC~\cite{sullivan2012overviewHEVC} and VVC~\cite{bross2021overviewVVC}, given their rising adoption in image and video compression within contemporary vision applications.

The contributions of this paper can be summarized as follows:
\begin{itemize}[leftmargin=*]
    \item We propose HybridAgent, a novel interactive paradigm for image restoration that integrates fast, slow, and feedback agents, enabling efficient and task-adaptive interactions to meet diverse user requirements and handle various degradations within a unified restoration framework. 
    \item We identify the error propagation issue associated with step-by-step single-distortion removal strategies and propose a three-stage training paradigm to construct restoration tools. This approach enhances the reuse of shared knowledge across tasks and introduces a mixed distortion removal pattern, effectively reducing restoration steps and mitigating error propagation in image restoration. 
    \item We build an instruction tuning dataset with over 100k image-text pairs across 10 distortion types to tailor our HybridAgent. The diversity of our dataset ensures the applicability of HybridAgent.
\end{itemize}
\begin{figure*}
    \centering
    \includegraphics[width=1\linewidth]{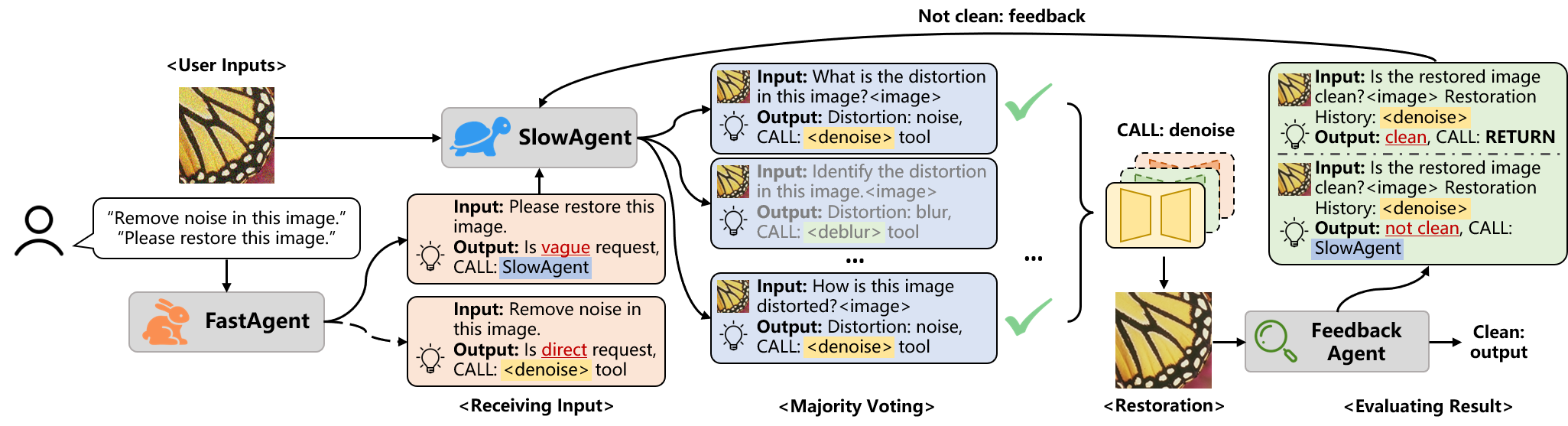}
    \caption{The overall pipeline of \textbf{HybridAgent}. We adopt a \textbf{FastAgent} to determine swiftly whether the user prompt is direct or vague. If a direct prompt is provided, \textbf{HybridAgent} will switch to the fast route (dashed lines) to invoke the corresponding restoration tool. Otherwise, \textbf{HybridAgent} will trigger the slow route (solid lines). \textbf{SlowAgent} automatically recognizes the distortion and executes the right restoration tool. To prevent incorrect tool invocation, we introduce a \textbf{FeedbackAgent} to assess whether the restored image is clean. \textbf{FeedbackAgent} and \textbf{SlowAgent} work collaboratively to generate the final clean output for the user.}
    \label{fig:agent}
\end{figure*}

\section{Related Works}
\label{sec:related}

\subsection{Image Restoration}
\noindent \textbf{Single-task Image Restoration.} Image restoration (IR)~\cite{arcnn,denoise1,denoise2,deblur1,dejpeg1,deblur3,denoise5} typically focuses on addressing a specific type of image degradation. Early works leverage convolutional neural networks (CNNs) due to their effective local information processing capabilities~\cite{arcnn,rdn,edsr,srcnn}. With the evolution of Transformers~\cite{vaswani2017attention}, which excel at uncovering long-range dependencies within the image, a series of transformer-based IR works~\cite{liang2021swinir,HAT,zamir2022restormer,wang2022uformer} have emerged. However, image restoration optimized with L1 or MSE loss often results in over-smoothed outputs~\cite{srgan}. To mitigate this, recent works introduce diffusion techniques~\cite{ddpm,ddim} into the IR network design. Benefiting from the powerful generation abilities, diffusion-based IR models~\cite{IR_diff1,xia2023diffir,ren2025moe,yu2024SUPIR,sun2024coser} can restore more realistic images with vivid texture details. 

Nevertheless, these works only focus on removing one type of distortion, which lacks adaptability towards various real-world application scenarios.

\noindent \textbf{All-in-one Image Restoration.} To address the above-mentioned problem, some works propose all-in-one IR models, which can handle multiple distortions with a unified model weight. Pioneer work AirNet~\cite{airnet} leverages contrastive learning to implicitly learn mappings between various degradation distributions and clean ones. Inspired by prompt learning in natural language processing (NLP), a series of works~\cite{potlapalli2023promptir,ma2023prores,PIP,ren2025moe,li2024ucip,da-clip} have studied prompt learning-based AIR, where a set of prompt parameters are implicitly encoded to capture various distortion representations. Meanwhile, due to the inherently multi-distorted nature of weather, several all-in-one approaches~\cite{weatherdiff,guo2024onerestore,li2020all} have emerged for removing weather-related distortions such as haze, rain, and snow. 

Although these approaches improve efficiency in handling various distortions, they lack control over distortion removal, \ie, they cannot selectively remove distortions according to user prompts. Recently, works such as InstructIR~\cite{conde2024instructIR} and UniProcessor~\cite{duan2025uniprocessor} leveraged text encoders to map user instructions as conditions of restoration models, enabling user-controllable AIR. However, these works lack generalization towards diverse user prompts due to limited encoding ability. Moreover, they adopt a step-by-step approach to handle hybrid distortions, which leads to suboptimal restoration results.

\subsection{Agent}

Agent typically refers to an intelligent system that can receive diverse user commands and automatically accomplish tasks accordingly~\cite{russell2016artificial}. Thanks to the great success of general-purposed large language models (LLMs)~\cite{brown2020languageGPT-3,roziere2023codellama,touvron2023llama2} and multimodal LLMs (MLLMs)~\cite{li2023blip,llava,openai2023gpt4}, such intelligent system can be realized through the combination of MLLMs and domain-specific expert models~\cite{schick2024toolformer,embodiedoctopus,mu2024embodiedgpt}. 

Despite extensive research on agents in high-level domains, how they can be effectively applied to low-level image restoration (IR) tasks remains an open question. As a pioneer work, RL-Restore~\cite{rl-restore} proposes to solve complex restoration problems in a sequential way. At each step, RL-Restore utilizes reinforcement learning to adaptively select the most suitable restoration tool for removing a specific type of distortion. Advanced by MLLMs, Clarity ChatGPT~\cite{claritychatgpt} enables dynamic tool selection by understanding the user's commands and performing reasoning. RestoreAgent~\cite{chen2024restoreagent} further optimizes the execution sequence of restoration tools and explores the most suitable model for specific degradation patterns. AgenticIR~\cite{agenticIR} improves the planning of the agent by incorporating proficient restoration experience, reflection, and rollback strategy.

Nevertheless, the aforementioned works all perform image restoration through step-by-step execution, neglecting the entanglement of distortions and potential distribution shift caused by different restoration models~\cite{li2020learning,chainofrestoration}. Moreover, they do not account for efficiency during tool invocation. How to adaptively and efficiently select restoration tools based on different user prompts remains an open research question.
\begin{figure*}
    \centering
    \includegraphics[width=1\linewidth]{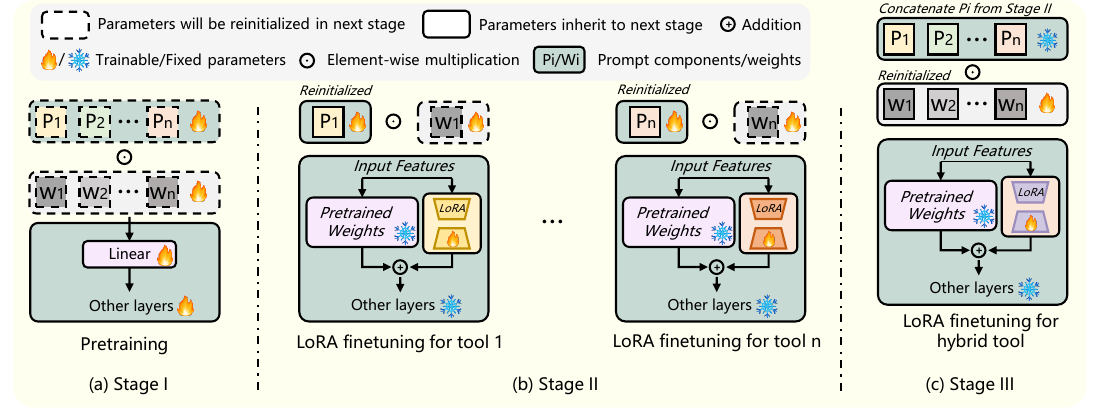}
    \caption{The illustration of three-stage training for the construction of restoration tools. We first build a well-trained base model following prompt learning-based all-in-one image restoration~\cite{potlapalli2023promptir,li2024promptcir} in Stage I. Subsequently, we build single-task restoration tools and hybrid restoration tool with LoRA~\cite{hu2021lora} in Stage II and III. Notably, we add LoRA to weights of \textit{Linear} layers in Attention modules and FeedForward modules of~\cite{li2024promptcir}. A more detailed diagram is provided in the \textbf{supplementary} due to limited space.}
    \label{fig:tools}
\end{figure*}

\section{Methods}
\label{sec:methods}

In this section, we introduce our HybridAgent, an intelligent agent system that dynamically and automatically handles complex IR problems. We discuss the details about how we realize the automatic image restoration through MLLM agents in Section~\ref{sec:hybridagent}. Specifically, to address the inefficiency of MLLM-based agents when handling simple user requests, we propose a collaborative agent system composed of FastAgent, SlowAgent, and FeedbackAgent, which efficiently responds to diverse user prompts and effectively solves complex IR tasks. Subsequently, we describe how we construct the restoration tools for HybridAgent in Section~\ref{sec:restoration}. Particularly, we propose a mixed distortion removal tool to proficiently address the error propagation problem in the step-by-step restoration process.

\subsection{HybridAgent}
\label{sec:hybridagent}

Imagine a user aiming to restore a degraded image. Typically, the user begins by identifying distortion types within the image and then searches for suitable restoration models. After applying a restoration step, the individual evaluates whether the image has been adequately restored. If degradations persist, the user continues to select and apply appropriate restoration tools iteratively until achieving a satisfactory restoration outcome. HybridAgent follows this same iterative procedure to automatically and efficiently restore degraded images, as shown in Figure~\ref{fig:agent}.

\noindent\textbf{Receiving Input.} HybridAgent receives a distorted image along with a user-provided prompt. Typically, an agent utilizes MLLMs to perceive the degradations and plan restoration steps automatically. However, professional users may provide explicit prompts clearly specifying their restoration requirements. In such cases, using MLLMs for degradation recognition becomes unnecessary, as the intended restoration type has already been identified. To handle this scenario efficiently, we employ the FastAgent based on a lightweight LLM. FastAgent directly analyzes the explicit user prompts, determines the required restoration tools, and proceeds immediately to the subsequent restoration step. On the contrary, if the prompt provided by the user is deemed ambiguous or unclear by FastAgent, the image will be handed over to SlowAgent and proceed to the next restoration step.

\noindent\textbf{Identifying Distortion.} If the image is handed over to SlowAgent, it is crucial to identify the proper distortion types and yield the suitable restoration tools. However, current MLLMs still face several challenges when handling such tasks: (i) existing MLLMs are not specifically fine-tuned for distortion type recognition, thus limiting their ability to accurately identify degradations, (ii) they cannot directly execute image restoration operations, and (iii) they lack the capability to determine whether further restoration steps are necessary.

To address the first challenge, we adopt an existing MLLM, Co-instruct~\cite{coinstruct}, which is specifically tuned for image quality assessment (IQA) and possesses distortion recognition capabilities. However, the degradation types covered by Co-instruct are insufficient for handling complex real-world requirements. Therefore, we further fine-tune Co-instruct on our proposed instruction-tuning dataset (detailed in the \textbf{supplementary}) to extend its distortion-recognition capabilities, resulting in the proposed SlowAgent. However, MLLMs may hallucinate and produce incorrect judgments. To mitigate this issue, inspired by the concept of test-time scaling~\cite{snell2024scaling}, we employ a majority voting mechanism that generates multiple candidate decisions and selects the most frequently occurring distortion as the final recognized distortion. To address the second challenge, we construct a set of restoration tools (detailed in Section~\ref{sec:restoration}) that can be invoked by SlowAgent. 

\noindent\textbf{Evaluating Result.} To address the third challenge, we establish a FeedbackAgent to assess the restored image. Such a FeedbackAgent is crucial because: (i) without external evaluation, SlowAgent cannot reliably determine if the restored image is already satisfactory; and (ii) it offers flexibility to terminate/continue the restoration process when a human user finds the intermediate result acceptable/unpleasant. Fortunately, determining whether a restored image is clean aligns closely with assessing image quality. However, leveraging IQA scores alone may not directly reflect whether an image meets restoration criteria. Thus, we propose the simplest way to provide feedback for SlowAgent: fine-tuning an IQA model (Co-Instruct) specifically to classify whether a restored image is clean. Nevertheless, achieving absolute perfection in restoration is nearly impossible in practice. Consequently, we further incorporate the historical information of the chosen restoration tools as the context for the FeedbackAgent, enabling it to more reliably determine whether the current image has reached a relatively clean state or still requires further restoration steps.

\subsection{Restoration Tools}
\label{sec:restoration}

After determining which tool should be invoked, the distorted image will be processed by the chosen restoration tool. Current tool designs~\cite{rl-restore,chen2024restoreagent,agenticIR} typically suffer from two primary limitations: (i) they directly adopt multiple single-task models trained separately, thus failing to effectively leverage common knowledge across different restoration tasks, and (ii) sequential application of these single-task models leads to step-by-step processing, which struggles to resolve distortion entanglement issues~\cite{li2020learning,chainofrestoration}. To simultaneously address these challenges, we propose a novel three-stage training strategy for restoration tools.

\noindent\textbf{Stage I.} In stage one, we aim to build a well-trained base model that shares common knowledge across various IR tasks. Inspired by the promising results of prompt learning-based methods~\cite{potlapalli2023promptir}, we adopt a similar network architecture. (We provide the details of this architecture in the \textbf{supplementary}.) To enhance the model's ability to represent diverse distortions (e.g., 10 types in this paper), we follow~\cite{li2024promptcir} and replace the first two stages' transformer blocks with shifted window attention blocks~\cite{liang2021swinir,HAT}.

\noindent\textbf{Stage II.} After obtaining the base model, we fine-tune task-specific models on different distortions to enable the agent to invoke the corresponding restoration tools. Once the base model is trained in stage I, it serves as a foundation model for distortion removal, allowing efficient fine-tuning to obtain the required restoration tools. Given LoRA’s~\cite{hu2021lora} effectiveness with minimal additional parameters, we employ it for fine-tuning in this stage, as shown in Figure~\ref{fig:tools}. Additionally, to ensure that the prompt accurately conveys the degradation condition to the model, we reinitialize the prompt parameters and fine-tune them jointly with LoRA parameters. Notably, we leverage prompts to encode descriptive information regarding distortions, while employing LoRA parameters to efficiently adapt the degradation-aware semantic information within deeper network structures.

\noindent\textbf{Stage III.} To address the challenge of mixed distortion removal in complex IR problems, we further tailor a mixed distortion removal tool. Similar to Stage II, we employ LoRA for efficient fine-tuning. However, we initialize the prompt parameters using those obtained in Stage II, enabling the network to effectively leverage both task-specific knowledge from Stage II and common knowledge from Stage I. We follow the distortion synthesis pipeline proposed in Real-ESRGAN~\cite{realesrgan} for training our model. More details are provided in the \textbf{supplementary}.

\begin{table*}[]

\caption{Comparison of proficiency and performance based on whether the fast route is enabled. ``A.I.T.'' represents ``Average Inference Time'', which indicates total inference time of agents. The inference time is evaluated on a RTX 4090D GPU. We use PSNR$\uparrow$/SSIM$\uparrow$ to evaluate the performance of two different settings.}
\resizebox{\textwidth}{!}{
\begin{tabular}{@{}c|cc|cc|cc|cc|cc@{}}
\toprule
 Setting & A.I.T. (s) & Performance & A.I.T. (s) & Performance & A.I.T. (s) & Performance & A.I.T. (s) & Performance & A.I.T. (s) & Performance \\ \midrule
  & \multicolumn{2}{c}{De-noise} & \multicolumn{2}{c}{De-blur} & \multicolumn{2}{c}{De-motionblur} & \multicolumn{2}{c}{De-jpeg} & \multicolumn{2}{c}{De-HEVC} \\ \midrule
  a) & 0.08 & 30.25/0.867 & 0.11 & 30.65/0.853 & 0.09 & 23.80/0.720 & 0.09 & 30.02/0.873 & 0.09 &  27.56/0.785 \\
  b) & 0.75 & 30.63/0.874 & 0.82 & 30.52/0.852 & 0.90 & 23.78/0.719 & 0.79 & 30.18/0.876 & 0.76 & 27.55/0.785 \\ \midrule\midrule
 & \multicolumn{2}{c}{De-VVC} & \multicolumn{2}{c}{De-rainstreak} & \multicolumn{2}{c}{De-raindrop} & \multicolumn{2}{c}{De-haze} & \multicolumn{2}{c}{De-low light} \\ \midrule
  a) & 0.09 & 27.91/0.797 & 0.13 & 30.04/0.893 & 0.12 & 30.30/0.913 & 0.09 & 29.92/0.960 & 0.12 & 22.60/0.825 \\
  b) & 0.79 & 27.92/0.798 & 1.05 & 30.03/0.893 & 0.94 & 30.34/0.914 & 0.83 & 29.92/0.960 & 0.88 & 22.61/0.828 \\ \bottomrule
\end{tabular}}
\label{tab:proficiency}
\end{table*}

\begin{table*}[!ht]
\caption{Comparisons of success rate between FastAgent and SlowAgent. As both agents will invoke tools, we define success rate as the proportion of the number of correct tool invocations to the total number of tool invocations.}
\resizebox{\textwidth}{!}{
\begin{tabular}{@{}l|cccccccccc@{}}
\toprule
 & De-noise & De-blur & De-motionblur & De-jpeg & De-HEVC & De-VVC & De-rainstreak & De-raindrop & De-haze & De-low light \\ \midrule
FastAgent & 72.9\% & 100.0\% & 100.0\% & 84.8\% & 96.2\% & 60.0\% & 100.0\% & 86.2\% & 100.0\% & 73.3\% \\
SlowAgent & 94.3\% & 94.3\% & 99.0\% & 96.7\% & 87.6\% & 90.9\% & 98.0\% & 96.5\% & 100.0\% & 100\% \\ \bottomrule
\end{tabular}}
\label{tab:acc}
\end{table*}

\section{Experiments}

\subsection{Implementation Details}
\noindent\textbf{Structure Details for HybridAgent.} Since most MLLMs are designed for general purposes, we fine-tune existing MLLMs to enable them to effectively serve the roles of SlowAgent and FeedbackAgent. We choose to fine-tune Co-Instruct~\cite{coinstruct} because it is capable of assessing distortions in images, providing an excellent starting point for our agents. As for restoration tools, we leverage an enhanced version of PromptIR~\cite{potlapalli2023promptir} from~\cite{li2024promptcir}, where the transformer blocks in the first two stages are replaced with RHAG~\cite{HAT} to improve the representative abilities. For the FastAgent, we adopt Llama3.2-1B-Instruct~\footnote{https://huggingface.co/meta-llama/Llama-3.2-1B-Instruct}.

\noindent\textbf{Training Datasets for Restoration Tools.} We optimize our restoration tools on 10 degradations. For noise, gaussian blur, motion blur, JPEG, HEVC, and VVC, we generate distorted samples online using 3450 images from DF2K~\cite{DIV2K,Flickr2K}. For other distortions, we adopt 1800 images from Rain100H~\cite{rain100H} for rainstreak, 861 images from RainDrop~\cite{raindrop} for raindrop, 6000 images from RESIDE-6k~\cite{reside-6k} for haze, and 485 images from LOL~\cite{lol} for low light following previous work~\cite{da-clip}. A more detailed explanation of degradation levels for synthesized images and the training details are given in the \textbf{supplementary}.

\noindent\textbf{Instruction Tuning Datasets for Agents.} Instruction tuning dataset plays an essential role in fine-tuning MLLM agents. Based on training datasets introduced above, we sample 5k images per distortion with resolution ranging between $224\times 224$ and $784\times 784$. We apply a linear transformation to map 5k to the total number of images within different distortion datasets, ensuring even sampling from each dataset. For hybrid distortion, we generate 20k images based on the combination of 10 distortions (more details are given in the \textbf{supplementary}). In total, our instruction tuning dataset for SlowAgent contains 70k image-text pairs. As for FeedbackAgent, we use images restored by the correct tools to synthesize 30k \textit{relatively} ``clean'' images and 33k ``not clean'' images by the incorrect tools. This results in a total of 66k image-text pairs for FeedbackAgent.

\noindent\textbf{Test Datasets.} Following~\cite{duan2025uniprocessor}, we adopt the combination of CBSD68~\cite{cbsd}, Urban100~\cite{urban100}, Kodak24~\cite{kodak}, and McMaster~\cite{mcmaster} to evaluate the performance on first six distortions. Following~\cite{duan2025uniprocessor}, we use 100 images from Rain100H~\cite{rain100H}, 58 images from RainDrop~\cite{raindrop}, 1000 images from RESIDE-6k~\cite{reside-6k}, and 15 images from LOL~\cite{lol} to evaluate the performance on last four distortions, respectively. We generate 200 images for mixed-degradation based on the mixture of 10 types of distortions. We provide more details including training settings in the \textbf{supplementary}.

\noindent\textbf{User Prompts.} We use GPT-4 to generate 20 direct textual prompts for each distortion type, along with an additional 20 ambiguous prompts, formulating a total of 220 diverse user prompts. We provide samples in the \textbf{supplementary}.

\begin{figure*}
    \centering
    \includegraphics[width=1\linewidth]{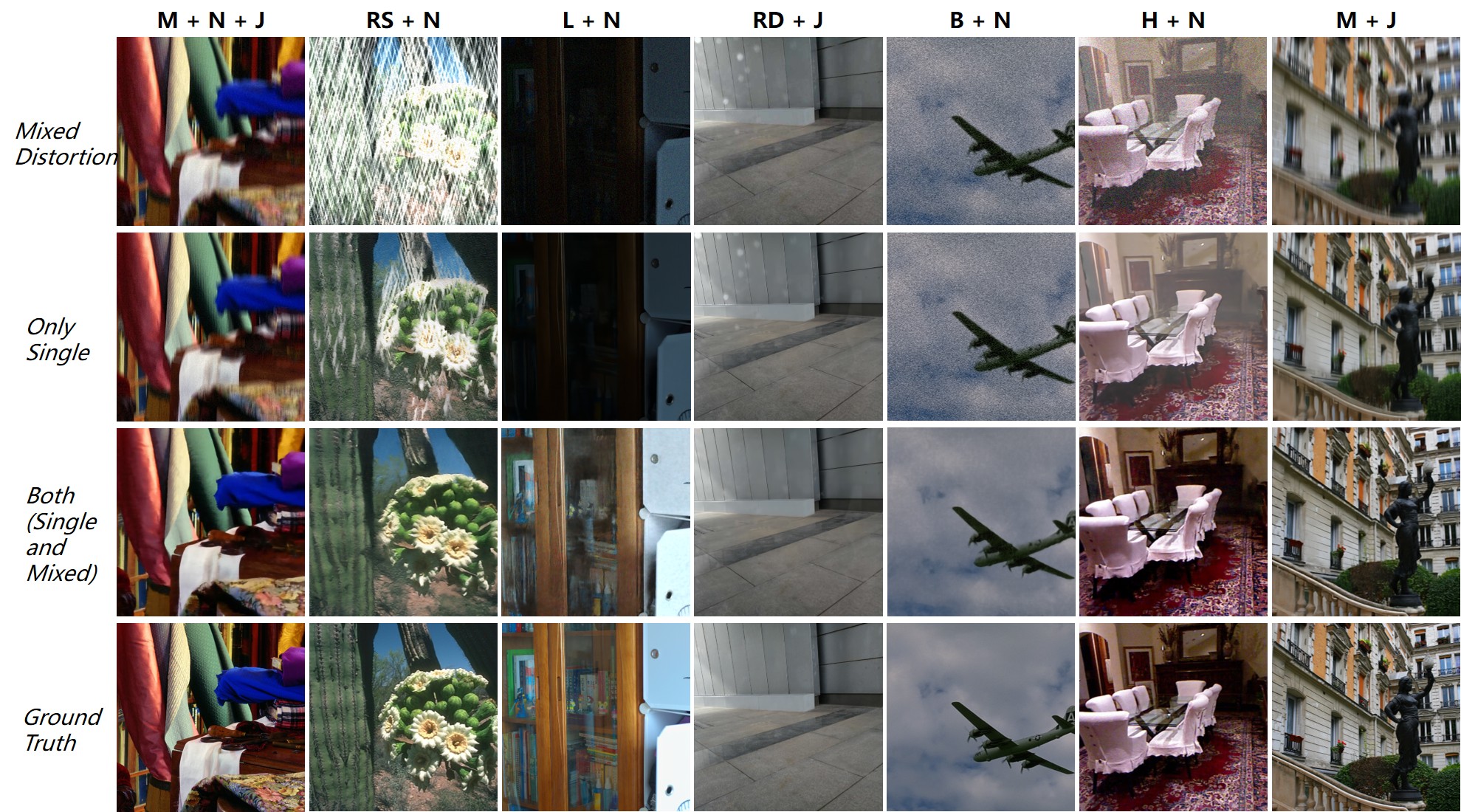}
    \caption{Qualitative comparisons of only single distortion removal tools against single and mixed distortion removal tools. \textbf{M}: Motionblur, \textbf{N}: Noise, \textbf{J}: JPEG, \textbf{RS}: Rainstreak, \textbf{L}: Low light, \textbf{RD}: Raindrop, \textbf{B}: Blur, \textbf{H}: Haze. Zoom in for a better view.}
    \label{fig:hyvsstep}
    \vspace{-3mm}
\end{figure*}

\subsection{Effectiveness of HybridAgent}
In this section, we evaluate the effectiveness of HybridAgent by addressing two key questions: i) Does FastAgent enhance the proficiency of the restoration pipeline? ii) Does mixed distortion removal outperform step-by-step distortion removal? Can they collaborate to address more complex distortions?

\subsubsection{Effectiveness of FastAgent Design} 
We compare two settings: a) the full HybridAgent, and b) HybridAgent with the fast route disabled, meaning that all images are processed through the SlowAgent, representing a traditional agent design. \textit{Notice that, to provide a more intuitive comparison, we use randomly selected direct user prompt in both settings.}  We evaluate the average inference time and performance on 10 distortions, as demonstrated in Table~\ref{tab:proficiency}. With the fast route enabled, HybridAgent achieves significantly higher operational efficiency for direct prompts (requiring only about 12\% of the runtime compared to SlowAgent), greatly surpassing the efficiency of SlowAgent. We further report the success rate of FastAgent and SlowAgent in Table~\ref{tab:acc}, defined as the proportion of the number of correct tool invocations to the total number of tool invocations. As observed, using in-context learning with FastAgent achieves a relatively high success rate, indicating that FastAgent can adapt to diverse user prompts in real-world scenarios and make accurate tool invocations.

\subsubsection{Single vs. Mixed Distortion Removal}
\label{sec:step-by-stepagainsthybridremoval}

As discussed in Section~\ref{sec:intro} and Section~\ref{sec:restoration}, removing mixed-degradation step-by-step may cause distribution shift and error propagation. We further validate this by comparing two settings: i) step-by-step restoration using only single distortion removal tools, and ii) restoration using both single and mixed distortion removal tools. As demonstrated in Table~\ref{tab:hybridstep}, mixed distortion removal significantly outperforms step-by-step single distortion removal across all metrics, particularly in handling haze and low-light distortions. This suggests that addressing mixed degradations with our proposed mixed distortion removal tools effectively mitigates error propagation and distribution shifts. We hypothesize that the poor performance of step-by-step removal for haze and low-light distortions is due to their unstable distortion modeling, which is easily disrupted by additional distortions. This is also evidenced in Figure~\ref{fig:hyvsstep}, where step-by-step restoration fails to enhance the low-light image or remove hazy artifacts.

\begin{figure}
    \centering
    \includegraphics[width=1\linewidth]{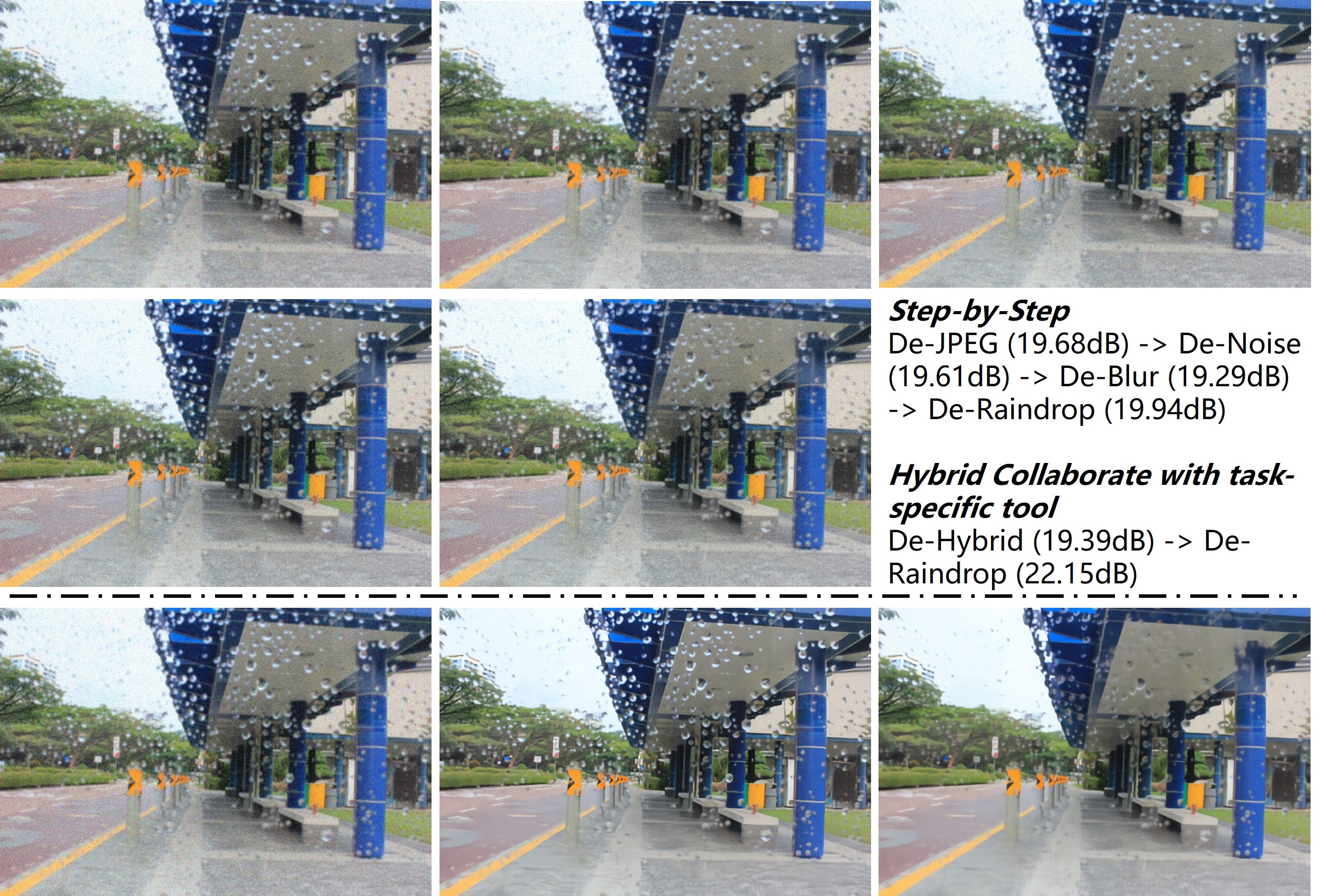}
    \caption{A case study on complex degradation removal. The image is corrupted by ``Raindrop + Blur + Noise + JPEG''. Upper: step-by-step distortion removal. Bottom: tools invoked by HybridAgent: De-hybrid + De-raindrop.}
    \label{fig:case}
    \vspace{-4.5mm}
\end{figure}

However, the representation ability of the mixed distortion removal tool is limited, constraining its applicability to more complex distortions or real-world scenarios. Nevertheless, the HybridAgent incorporates the FeedbackAgent, allowing the mixed distortion removal tool to be utilized in step-by-step scheduling. Consequently, the mixed distortion removal tool and single distortion removal tools can collaboratively address more complex distortion scenarios, reducing the step of tool invocations and improving overall performance. We provide a case study in Figure~\ref{fig:case}. The collaboration between the mixed distortion and single distortion removal tools reduces distribution shifts and achieves better restoration results in fewer steps.

\begin{table*}[]
\caption{Comparison of removing mixed distortions between only using single distortion removal tools and using both single and mixed distortion removal tools. }
\vspace{-2mm}
\resizebox{\textwidth}{!}{
\begin{tabular}{@{}c|ccc|ccc|ccc|ccc|ccc@{}}
\toprule
 & \multicolumn{3}{c|}{Blur + Noise} & \multicolumn{3}{c|}{Blur + JPEG} & \multicolumn{3}{c|}{Blur + Noise + JPEG} & \multicolumn{3}{c|}{Motionblur + Noise} & \multicolumn{3}{c}{Motionblur + JPEG} \\ \cmidrule(l){2-16} 
 & PSNR$\uparrow$ & SSIM$\uparrow$ & LPIPS$\downarrow$ & PSNR$\uparrow$ & SSIM$\uparrow$ & LPIPS$\downarrow$ & PSNR$\uparrow$ & SSIM$\uparrow$ & LPIPS$\downarrow$ & PSNR$\uparrow$ & SSIM$\uparrow$ & LPIPS$\downarrow$ & PSNR$\uparrow$ & SSIM$\uparrow$ & LPIPS$\downarrow$ \\ \midrule
Only Single & 23.72 & 0.555 & 0.520 & 26.04 & 0.737 & 0.300 & 22.32 & 0.423 & 0.640 & 22.10 & 0.532 & 0.501 & 23.79 & 0.668 & 0.360 \\
Both & 26.21 & 0.733 & 0.311 & 26.54 & 0.775 & 0.278 & 25.37 & 0.706 & 0.352 & 23.13 & 0.628 & 0.388 & 23.77 & 0.674 & 0.330  \\ \midrule\midrule
& \multicolumn{3}{c|}{Motionblur + Noise + JPEG} & \multicolumn{3}{c|}{Rainstreak + Noise} & \multicolumn{3}{c|}{Rainstreak + JPEG} & \multicolumn{3}{c|}{Raindrop + Noise} & \multicolumn{3}{c}{Raindrop + JPEG} \\ \cmidrule(l){2-16} 
 & PSNR$\uparrow$ & SSIM$\uparrow$ & LPIPS$\downarrow$ & PSNR$\uparrow$ & SSIM$\uparrow$ & LPIPS$\downarrow$ & PSNR$\uparrow$ & SSIM$\uparrow$ & LPIPS$\downarrow$ & PSNR$\uparrow$ & SSIM$\uparrow$ & LPIPS$\downarrow$ & PSNR$\uparrow$ & SSIM$\uparrow$ & LPIPS$\downarrow$ \\ \midrule
Only Single & 20.66 & 0.439 & 0.551 & 23.36 & 0.622 & 0.351 & 22.18 & 0.692 & 0.300 & 26.51 & 0.743 & 0.287 & 25.40 & 0.814 & 0.233 \\
Both & 22.82 & 0.618 & 0.404 & 26.49 & 0.766 & 0.203 & 25.44 & 0.764 & 0.242 & 27.98 & 0.807 & 0.191 & 28.42 & 0.835 & 0.188 \\ \midrule\midrule
& \multicolumn{3}{c|}{Haze + Noise} & \multicolumn{3}{c|}{Haze + JPEG} & \multicolumn{3}{c|}{Low light + Noise} & \multicolumn{3}{c|}{Low light + JPEG} & \multicolumn{3}{c}{Average} \\ \cmidrule(l){2-16} 
 & PSNR$\uparrow$ & SSIM$\uparrow$ & LPIPS$\downarrow$ & PSNR$\uparrow$ & SSIM$\uparrow$ & LPIPS$\downarrow$ & PSNR$\uparrow$ & SSIM$\uparrow$ & LPIPS$\downarrow$ & PSNR$\uparrow$ & SSIM$\uparrow$ & LPIPS$\downarrow$ & PSNR$\uparrow$ & SSIM$\uparrow$ & LPIPS$\downarrow$ \\ \midrule
Only Single & 14.42 & 0.536 & 0.497 & 12.57 & 0.668 & 0.302 & 7.69 & 0.188 & 0.761 & 7.05 & 0.174 & 0.616 & 19.84 & 0.557 & 0.444 \\
Both & 23.40 & 0.797 & 0.184 & 26.43 & 0.885 & 0.113 & 19.97 & 0.627 & 0.475 & 21.65 & 0.754 & 0.324 & 24.83 & 0.741 & 0.284 \\ \bottomrule
\end{tabular}}
\label{tab:hybridstep}
\vspace{-2mm}
\end{table*}

\begin{table*}[]
% \vspace{3.2mm}
\caption{Comparison of HybridAgent with All-in-one image restoration methods. For SwinIR, Uformer, Restormer, and PromptIR, we retrain them following their official code on our 10 degradation datasets. For InstructIR, we only test on the seen distortions for a fair comparison. Best performances are \textbf{bolded}.}
\vspace{-1mm}
\resizebox{\textwidth}{!}{
\begin{tabular}{@{}l|ccc|ccc|ccc|ccc|ccc@{}}
\toprule
 & \multicolumn{3}{c|}{Rainstreak + Noise} & \multicolumn{3}{c|}{Haze + Noise} & \multicolumn{3}{c|}{Low light + Noise} & \multicolumn{3}{c|}{Motionblur + Noise} & \multicolumn{3}{c}{Motionblur + Noise + JPEG} \\ \midrule
 & PSNR$\uparrow$ & SSIM$\uparrow$ & LPIPS$\downarrow$ & PSNR$\uparrow$ & SSIM$\uparrow$ & LPIPS$\downarrow$ & PSNR$\uparrow$ & SSIM$\uparrow$ & LPIPS$\downarrow$ & PSNR$\uparrow$ & SSIM$\uparrow$ & LPIPS$\downarrow$ & PSNR$\uparrow$ & SSIM$\uparrow$ & LPIPS$\downarrow$ \\ \midrule
SwinIR~\cite{liang2021swinir} & 20.83 & 0.594 & 0.429 & 15.04 & 0.545 & 0.504 & 7.52 & 0.168 & 0.839 & 20.98 & 0.491 & 0.560 & 19.65 & 0.391 & 0.646 \\
Uformer~\cite{wang2022uformer} & 21.37 & 0.565 & 0.409 & 16.27 & 0.587 & 0.452 & 9.28 & 0.289 & 0.781 & 21.17 & 0.494 & 0.549 & 19.84 & 0.381 & 0.646 \\
Restormer~\cite{zamir2022restormer} & 23.57 & 0.605 & 0.377 & 15.37 & 0.538 & 0.500 & 15.07 & 0.546 & 0.603 & 20.72 & 0.499 &  0.544& 19.82 & 0.364 & 0.640 \\
AirNet~\cite{airnet} & 22.95 & 0.588 & 0.393 & 14.33 & 0.510 & 0.544 & 15.72 & 0.538 & 0.612 & 20.68 & 0.487 & 0.561 & 19.29 & 0.340 & 0.673 \\
PromptIR~\cite{potlapalli2023promptir} & 23.17 & 0.591 & 0.388 & 14.54 & 0.513 & 0.532 & 15.87 & 0.541 & 0.610 & 20.79 & 0.489 & 0.558 & 19.40 & 0.342 & 0.668 \\
InstructIR~\cite{conde2024instructIR} & 14.63 & 0.431 & 0.534 & 18.99 & 0.615 & 0.426 & 19.15 & 0.599 & 0.526 & 20.99 & 0.493 & 0.542 & - & - & - \\
HybridAgent & \textbf{26.49} & \textbf{0.766} & \textbf{0.203} & \textbf{23.40} & \textbf{0.797} & \textbf{0.184} & \textbf{19.97}& \textbf{0.627} & \textbf{0.475} & \textbf{23.77} & \textbf{0.674} & \textbf{0.330} & \textbf{22.82} & \textbf{0.618} & \textbf{0.404} \\ \bottomrule
\end{tabular}}
\label{tab:sota}
\vspace{-3mm}
\end{table*}

\subsection{Comparisons with All-in-One Methods}
\begin{figure}
    \centering
    \includegraphics[width=1\linewidth]{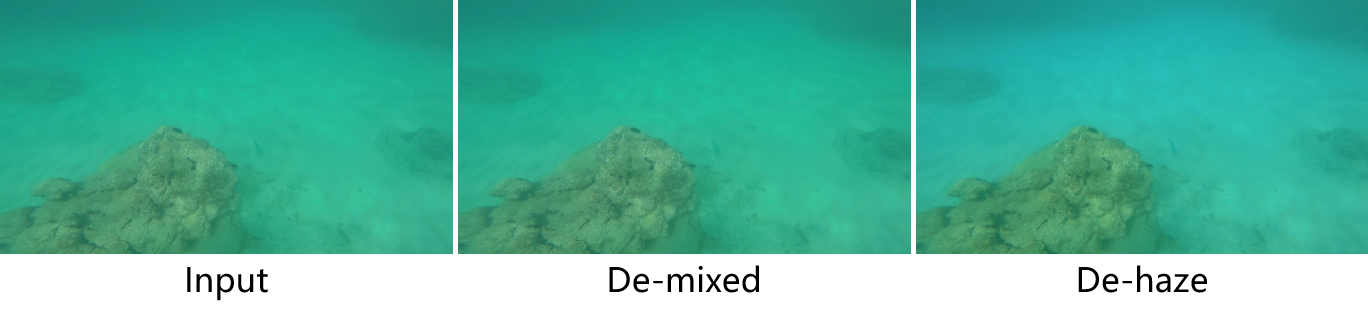}
    \caption{Visualization of the restoration process performed by HybridAgent on real-world image (test9003up.jpg from EUVP~\cite{underwater} dataset). Zoom in for best views.}
    \label{fig:realworld}
    \vspace{-5mm}
\end{figure}

To further demonstrate the effectiveness of HybridAgent against state-of-the-art all-in-one IR methods, we evaluate the performance on complex restoration scenarios, including both synthetic and real-world datasets. For a fair comparison, we have retrained SwinIR~\cite{liang2021swinir}, Uformer~\cite{wang2022uformer}, AirNet~\cite{airnet}, Restormer~\cite{zamir2022restormer}, and PromptIR~\cite{potlapalli2023promptir} on our proposed 10 degradation datasets. Following~\cite{chen2024restoreagent}, \textit{we infer each of the above methods multiple times in a sequential way to achieve the best performance.} For InstructIR~\cite{conde2024instructIR}, we test the method on seen distortions, ensuring a fair comparison with HybridAgent. As demonstrated in Table~\ref{tab:sota}, our method achieves significant improvements against other all-in-one methods on mixed degradations. Compared to traditional all-in-one methods, such as PromptIR, HybridAgent benefits from task-specific tools and the mixed distortion removal tool, which enjoys not only common knowledge across various distortions but also task-specific expertise. Additionally, InstructIR executes the restoration process following human instructions. However, this implementation may cause error propagation during the restoration process, resulting in suboptimal outputs. Conversely, HybridAgent incorporates the mixed distortion removal tool, which efficiently mitigates the impact of incorrect restoration decisions. For qualitative comparisons, we provide visual results on real-world underwater unpaired dataset EUVP~\cite{underwater} in Figure~\ref{fig:realworld}. As observed, HybridAgent first takes mixed distortion removal tool to remove entangled degradations, and then utilizes dehaze tool to further enhance the visual quality of the image. We provide more qualitative results on synthetic datasets and other ablation studies about our three-stage training design of restoration tools in the \textbf{supplementary} due to limited space.

% \vspace{3.2mm}

\section{Conclusion}

In this work, we present HybridAgents by introducing the composition of fast and slow restoration agents, intending to solve two challenges in existing agent/instruction-based image restoration works: (i) lacking flexibility on task difficulty; and (ii) the error propagation since the step-by-step single restoration tool invoking. To tackle the first challenge, we introduce hybrid restoration agents, where FastAgent is responsible for simple and clear user requirements while SlowAgent is optimized with our proposed large instruction tuning dataset to support ambiguous user requirements. A FeedbackAgent is also designed to collaborate with SlowAgent to provide accurate feedback and perform the termination of the restoration process. For the second challenge, we propose a three-stage training strategy which introduces the abundant mixed distortion removal tool and enhances the model reusing capability by optimizing the model with multi-task learning and task-specific prompt optimization. Extensive experiments on broad user requirements and various complicated degradations have demonstrated the effectiveness of our HybridAgent. 

%%%%%
{
    \small
    \bibliographystyle{ieeenat_fullname}
    \bibliography{main}

\begin{thebibliography}{87}
\providecommand{\natexlab}[1]{#1}
\providecommand{\url}[1]{\texttt{#1}}
\expandafter\ifx\csname urlstyle\endcsname\relax
  \providecommand{\doi}[1]{doi: #1}\else
  \providecommand{\doi}{doi: \begingroup \urlstyle{rm}\Url}\fi

\bibitem[Agustsson and Timofte(2017)]{DIV2K}
Eirikur Agustsson and Radu Timofte.
\newblock Ntire 2017 challenge on single image super-resolution: Dataset and study.
\newblock In \emph{Proceedings of the IEEE conference on computer vision and pattern recognition workshops}, pages 126--135, 2017.

\bibitem[Bai et~al.(2023)Bai, Bai, Yang, Wang, Tan, Wang, Lin, Zhou, and Zhou]{bai2023qwen}
Jinze Bai, Shuai Bai, Shusheng Yang, Shijie Wang, Sinan Tan, Peng Wang, Junyang Lin, Chang Zhou, and Jingren Zhou.
\newblock Qwen-vl: A frontier large vision-language model with versatile abilities.
\newblock \emph{arXiv preprint arXiv:2308.12966}, 2023.

\bibitem[Bross et~al.(2021{\natexlab{a}})Bross, Wang, Ye, Liu, Chen, Sullivan, and Ohm]{VTM}
Benjamin Bross, Ye-Kui Wang, Yan Ye, Shan Liu, Jianle Chen, Gary~J Sullivan, and Jens-Rainer Ohm.
\newblock Overview of the versatile video coding (vvc) standard and its applications.
\newblock \emph{IEEE Transactions on Circuits and Systems for Video Technology}, 31\penalty0 (10):\penalty0 3736--3764, 2021{\natexlab{a}}.

\bibitem[Bross et~al.(2021{\natexlab{b}})Bross, Wang, Ye, Liu, Chen, Sullivan, and Ohm]{bross2021overviewVVC}
Benjamin Bross, Ye-Kui Wang, Yan Ye, Shan Liu, Jianle Chen, Gary~J Sullivan, and Jens-Rainer Ohm.
\newblock Overview of the versatile video coding (vvc) standard and its applications.
\newblock \emph{IEEE Transactions on Circuits and Systems for Video Technology}, 31\penalty0 (10):\penalty0 3736--3764, 2021{\natexlab{b}}.

\bibitem[Brown et~al.(2020)Brown, Mann, Ryder, Subbiah, Kaplan, Dhariwal, Neelakantan, Shyam, Sastry, Askell, et~al.]{brown2020languageGPT-3}
Tom Brown, Benjamin Mann, Nick Ryder, Melanie Subbiah, Jared~D Kaplan, Prafulla Dhariwal, Arvind Neelakantan, Pranav Shyam, Girish Sastry, Amanda Askell, et~al.
\newblock Language models are few-shot learners.
\newblock \emph{Advances in neural information processing systems}, 33:\penalty0 1877--1901, 2020.

\bibitem[Cao et~al.(2024)Cao, Meng, and Cao]{chainofrestoration}
Jin Cao, Deyu Meng, and Xiangyong Cao.
\newblock Chain-of-restoration: Multi-task image restoration models are zero-shot step-by-step universal image restorers.
\newblock \emph{arXiv preprint arXiv:2410.08688}, 2024.

\bibitem[Chen et~al.(2024)Chen, Li, Gu, Ren, Chen, Ye, Pei, Zhou, Song, and Zhu]{chen2024restoreagent}
Haoyu Chen, Wenbo Li, Jinjin Gu, Jingjing Ren, Sixiang Chen, Tian Ye, Renjing Pei, Kaiwen Zhou, Fenglong Song, and Lei Zhu.
\newblock Restoreagent: Autonomous image restoration agent via multimodal large language models.
\newblock \emph{arXiv preprint arXiv:2407.18035}, 2024.

\bibitem[Chen et~al.(2023{\natexlab{a}})Chen, Wang, Zhou, Qiao, and Dong]{HAT}
Xiangyu Chen, Xintao Wang, Jiantao Zhou, Yu Qiao, and Chao Dong.
\newblock Activating more pixels in image super-resolution transformer.
\newblock In \emph{Proceedings of the IEEE/CVF conference on computer vision and pattern recognition}, pages 22367--22377, 2023{\natexlab{a}}.

\bibitem[Chen et~al.(2023{\natexlab{b}})Chen, Wu, Wang, Su, Chen, Xing, Muyan, Zhang, Zhu, Lu, et~al.]{chen2023internvl}
Zhe Chen, Jiannan Wu, Wenhai Wang, Weijie Su, Guo Chen, Sen Xing, Zhong Muyan, Qinglong Zhang, Xizhou Zhu, Lewei Lu, et~al.
\newblock Internvl: Scaling up vision foundation models and aligning for generic visual-linguistic tasks.
\newblock \emph{arXiv preprint arXiv:2312.14238}, 2023{\natexlab{b}}.

\bibitem[Conde et~al.(2024)Conde, Geigle, and Timofte]{conde2024instructIR}
Marcos~V Conde, Gregor Geigle, and Radu Timofte.
\newblock High-quality image restoration following human instructions.
\newblock \emph{arXiv preprint arXiv:2401.16468}, 2024.

\bibitem[Dong et~al.(2015{\natexlab{a}})Dong, Deng, Loy, and Tang]{arcnn}
Chao Dong, Yubin Deng, Chen~Change Loy, and Xiaoou Tang.
\newblock Compression artifacts reduction by a deep convolutional network.
\newblock In \emph{Proceedings of the IEEE international conference on computer vision}, pages 576--584, 2015{\natexlab{a}}.

\bibitem[Dong et~al.(2015{\natexlab{b}})Dong, Loy, He, and Tang]{srcnn}
Chao Dong, Chen~Change Loy, Kaiming He, and Xiaoou Tang.
\newblock Image super-resolution using deep convolutional networks.
\newblock \emph{IEEE transactions on pattern analysis and machine intelligence}, 38\penalty0 (2):\penalty0 295--307, 2015{\natexlab{b}}.

\bibitem[Duan et~al.(2025)Duan, Min, Wu, Shen, and Zhai]{duan2025uniprocessor}
Huiyu Duan, Xiongkuo Min, Sijing Wu, Wei Shen, and Guangtao Zhai.
\newblock Uniprocessor: a text-induced unified low-level image processor.
\newblock In \emph{European Conference on Computer Vision}, pages 180--199. Springer, 2025.

\bibitem[Engin et~al.(2018)Engin, Gen{\c{c}}, and Kemal~Ekenel]{engin2018cycledehaze}
Deniz Engin, Anil Gen{\c{c}}, and Hazim Kemal~Ekenel.
\newblock Cycle-dehaze: Enhanced cyclegan for single image dehazing.
\newblock In \emph{Proceedings of the IEEE conference on computer vision and pattern recognition workshops}, pages 825--833, 2018.

\bibitem[Fan et~al.(2019)Fan, Zhang, Fan, and Zhang]{denoise1}
Linwei Fan, Fan Zhang, Hui Fan, and Caiming Zhang.
\newblock Brief review of image denoising techniques.
\newblock \emph{Visual Computing for Industry, Biomedicine, and Art}, 2\penalty0 (1):\penalty0 7, 2019.

\bibitem[Fei et~al.(2023)Fei, Lyu, Pan, Zhang, Yang, Luo, Zhang, and Dai]{IR_diff1}
Ben Fei, Zhaoyang Lyu, Liang Pan, Junzhe Zhang, Weidong Yang, Tianyue Luo, Bo Zhang, and Bo Dai.
\newblock Generative diffusion prior for unified image restoration and enhancement.
\newblock In \emph{Proceedings of the IEEE/CVF Conference on Computer Vision and Pattern Recognition}, pages 9935--9946, 2023.

\bibitem[Franzen(1999)]{kodak}
Rich Franzen.
\newblock Kodak lossless true color image suite.
\newblock \url{http://r0k.us/graphics/kodak/}, 1999.
\newblock Online accessed 24 Oct 2021.

\bibitem[Fu et~al.(2021)Fu, Wang, Liu, Han, and Zha]{dejpeg2}
Xueyang Fu, Xi Wang, Aiping Liu, Junwei Han, and Zheng-Jun Zha.
\newblock Learning dual priors for jpeg compression artifacts removal.
\newblock In \emph{Proceedings of the IEEE/CVF international conference on computer vision}, pages 4086--4095, 2021.

\bibitem[Galteri et~al.(2017)Galteri, Seidenari, Bertini, and Del~Bimbo]{dejpeg1}
Leonardo Galteri, Lorenzo Seidenari, Marco Bertini, and Alberto Del~Bimbo.
\newblock Deep generative adversarial compression artifact removal.
\newblock In \emph{Proceedings of the IEEE international conference on computer vision}, pages 4826--4835, 2017.

\bibitem[Guo et~al.(2024)Guo, Gao, Lu, Zhu, Liu, and He]{guo2024onerestore}
Yu Guo, Yuan Gao, Yuxu Lu, Huilin Zhu, Ryan~Wen Liu, and Shengfeng He.
\newblock Onerestore: A universal restoration framework for composite degradation.
\newblock \emph{arXiv preprint arXiv:2407.04621}, 2024.

\bibitem[Ho et~al.(2020)Ho, Jain, and Abbeel]{ddpm}
Jonathan Ho, Ajay Jain, and Pieter Abbeel.
\newblock Denoising diffusion probabilistic models.
\newblock \emph{Advances in neural information processing systems}, 33:\penalty0 6840--6851, 2020.

\bibitem[Hu et~al.(2021)Hu, Shen, Wallis, Allen-Zhu, Li, Wang, Wang, and Chen]{hu2021lora}
Edward~J Hu, Yelong Shen, Phillip Wallis, Zeyuan Allen-Zhu, Yuanzhi Li, Shean Wang, Lu Wang, and Weizhu Chen.
\newblock Lora: Low-rank adaptation of large language models.
\newblock \emph{arXiv preprint arXiv:2106.09685}, 2021.

\bibitem[Huang et~al.(2015)Huang, Singh, and Ahuja]{urban100}
Jia-Bin Huang, Abhishek Singh, and Narendra Ahuja.
\newblock Single image super-resolution from transformed self-exemplars.
\newblock In \emph{Proceedings of the IEEE conference on computer vision and pattern recognition}, pages 5197--5206, 2015.

\bibitem[Islam et~al.(2020)Islam, Xia, and Sattar]{underwater}
Md~Jahidul Islam, Youya Xia, and Junaed Sattar.
\newblock Fast underwater image enhancement for improved visual perception.
\newblock \emph{IEEE Robotics and Automation Letters}, 5\penalty0 (2):\penalty0 3227--3234, 2020.

\bibitem[Jiang et~al.(2021)Jiang, Zhang, and Timofte]{dejpeg3}
Jiaxi Jiang, Kai Zhang, and Radu Timofte.
\newblock Towards flexible blind jpeg artifacts removal.
\newblock In \emph{Proceedings of the IEEE/CVF International Conference on Computer Vision}, pages 4997--5006, 2021.

\bibitem[Ledig et~al.(2017)Ledig, Theis, Husz{\'a}r, Caballero, Cunningham, Acosta, Aitken, Tejani, Totz, Wang, et~al.]{srgan}
Christian Ledig, Lucas Theis, Ferenc Husz{\'a}r, Jose Caballero, Andrew Cunningham, Alejandro Acosta, Andrew Aitken, Alykhan Tejani, Johannes Totz, Zehan Wang, et~al.
\newblock Photo-realistic single image super-resolution using a generative adversarial network.
\newblock In \emph{Proceedings of the IEEE conference on computer vision and pattern recognition}, pages 4681--4690, 2017.

\bibitem[Li et~al.(2022)Li, Liu, Hu, Wu, Lv, and Peng]{airnet}
Boyun Li, Xiao Liu, Peng Hu, Zhongqin Wu, Jiancheng Lv, and Xi Peng.
\newblock All-in-one image restoration for unknown corruption.
\newblock In \emph{Proceedings of the IEEE/CVF Conference on Computer Vision and Pattern Recognition}, pages 17452--17462, 2022.

\bibitem[Li et~al.(2024{\natexlab{a}})Li, Li, Lu, Feng, Guo, Zhao, Zhang, and Chen]{li2024promptcir}
Bingchen Li, Xin Li, Yiting Lu, Ruoyu Feng, Mengxi Guo, Shijie Zhao, Li Zhang, and Zhibo Chen.
\newblock Promptcir: Blind compressed image restoration with prompt learning.
\newblock \emph{arXiv preprint arXiv:2404.17433}, 2024{\natexlab{a}}.

\bibitem[Li et~al.(2023{\natexlab{a}})Li, Li, Savarese, and Hoi]{li2023blip}
Junnan Li, Dongxu Li, Silvio Savarese, and Steven Hoi.
\newblock Blip-2: Bootstrapping language-image pre-training with frozen image encoders and large language models.
\newblock In \emph{International conference on machine learning}, pages 19730--19742. PMLR, 2023{\natexlab{a}}.

\bibitem[Li et~al.(2020{\natexlab{a}})Li, Tan, and Cheong]{li2020all}
Ruoteng Li, Robby~T Tan, and Loong-Fah Cheong.
\newblock All in one bad weather removal using architectural search.
\newblock In \emph{Proceedings of the IEEE/CVF conference on computer vision and pattern recognition}, pages 3175--3185, 2020{\natexlab{a}}.

\bibitem[Li et~al.(2020{\natexlab{b}})Li, Jin, Lin, Liu, Wu, Yu, Zhou, and Chen]{li2020learning}
Xin Li, Xin Jin, Jianxin Lin, Sen Liu, Yaojun Wu, Tao Yu, Wei Zhou, and Zhibo Chen.
\newblock Learning disentangled feature representation for hybrid-distorted image restoration.
\newblock In \emph{Computer Vision--ECCV 2020: 16th European Conference, Glasgow, UK, August 23--28, 2020, Proceedings, Part XXIX 16}, pages 313--329. Springer, 2020{\natexlab{b}}.

\bibitem[Li et~al.(2024{\natexlab{b}})Li, Li, Jin, Lan, Zhu, Ren, and Chen]{li2024ucip}
Xin Li, Bingchen Li, Yeying Jin, Cuiling Lan, Hanxin Zhu, Yulin Ren, and Zhibo Chen.
\newblock Ucip: A universal framework for compressed image super-resolution using dynamic prompt.
\newblock \emph{arXiv preprint arXiv:2407.13108}, 2024{\natexlab{b}}.

\bibitem[Li et~al.(2023{\natexlab{b}})Li, Fan, Xiang, Demandolx, Ranjan, Timofte, and Van~Gool]{li2023efficientIR}
Yawei Li, Yuchen Fan, Xiaoyu Xiang, Denis Demandolx, Rakesh Ranjan, Radu Timofte, and Luc Van~Gool.
\newblock Efficient and explicit modelling of image hierarchies for image restoration.
\newblock In \emph{Proceedings of the IEEE/CVF Conference on Computer Vision and Pattern Recognition}, pages 18278--18289, 2023{\natexlab{b}}.

\bibitem[Li et~al.(2023{\natexlab{c}})Li, Lei, Ma, Zhang, and Shan]{PIP}
Zilong Li, Yiming Lei, Chenglong Ma, Junping Zhang, and Hongming Shan.
\newblock Prompt-in-prompt learning for universal image restoration.
\newblock \emph{arXiv preprint arXiv:2312.05038}, 2023{\natexlab{c}}.

\bibitem[Liang et~al.(2021)Liang, Cao, Sun, Zhang, Van~Gool, and Timofte]{liang2021swinir}
Jingyun Liang, Jiezhang Cao, Guolei Sun, Kai Zhang, Luc Van~Gool, and Radu Timofte.
\newblock Swinir: Image restoration using swin transformer.
\newblock In \emph{Proceedings of the IEEE/CVF international conference on computer vision}, pages 1833--1844, 2021.

\bibitem[Lim et~al.(2017)Lim, Son, Kim, Nah, and Mu~Lee]{edsr}
Bee Lim, Sanghyun Son, Heewon Kim, Seungjun Nah, and Kyoung Mu~Lee.
\newblock Enhanced deep residual networks for single image super-resolution.
\newblock In \emph{Proceedings of the IEEE conference on computer vision and pattern recognition workshops}, pages 136--144, 2017.

\bibitem[Liu et~al.(2024)Liu, Li, Wu, and Lee]{llava}
Haotian Liu, Chunyuan Li, Qingyang Wu, and Yong~Jae Lee.
\newblock Visual instruction tuning.
\newblock \emph{Advances in neural information processing systems}, 36, 2024.

\bibitem[Loshchilov and Hutter(2016)]{loshchilov2016sgdr}
Ilya Loshchilov and Frank Hutter.
\newblock Sgdr: Stochastic gradient descent with warm restarts.
\newblock \emph{arXiv preprint arXiv:1608.03983}, 2016.

\bibitem[Luo et~al.(2023)Luo, Gustafsson, Zhao, Sj{\"o}lund, and Sch{\"o}n]{da-clip}
Ziwei Luo, Fredrik~K Gustafsson, Zheng Zhao, Jens Sj{\"o}lund, and Thomas~B Sch{\"o}n.
\newblock Controlling vision-language models for universal image restoration.
\newblock \emph{arXiv preprint arXiv:2310.01018}, 2023.

\bibitem[Ma et~al.(2023)Ma, Cheng, Wang, Zhang, Wang, and Zhang]{ma2023prores}
Jiaqi Ma, Tianheng Cheng, Guoli Wang, Qian Zhang, Xinggang Wang, and Lefei Zhang.
\newblock Prores: Exploring degradation-aware visual prompt for universal image restoration.
\newblock \emph{arXiv preprint arXiv:2306.13653}, 2023.

\bibitem[Martin et~al.(2001)Martin, Fowlkes, Tal, and Malik]{cbsd}
David Martin, Charless Fowlkes, Doron Tal, and Jitendra Malik.
\newblock A database of human segmented natural images and its application to evaluating segmentation algorithms and measuring ecological statistics.
\newblock In \emph{Proceedings eighth IEEE international conference on computer vision. ICCV 2001}, pages 416--423. IEEE, 2001.

\bibitem[Mu et~al.(2024)Mu, Zhang, Hu, Wang, Ding, Jin, Wang, Dai, Qiao, and Luo]{mu2024embodiedgpt}
Yao Mu, Qinglong Zhang, Mengkang Hu, Wenhai Wang, Mingyu Ding, Jun Jin, Bin Wang, Jifeng Dai, Yu Qiao, and Ping Luo.
\newblock Embodiedgpt: Vision-language pre-training via embodied chain of thought.
\newblock \emph{Advances in Neural Information Processing Systems}, 36, 2024.

\bibitem[Nah et~al.(2021)Nah, Son, Lee, Timofte, Lee, Chen, Zhang, Lu, Chu, Chen, et~al.]{deblur1}
Seungjun Nah, Sanghyun Son, Suyoung Lee, Radu Timofte, Kyoung~Mu Lee, Liangyu Chen, Jie Zhang, Xin Lu, Xiaojie Chu, Chengpeng Chen, et~al.
\newblock Ntire 2021 challenge on image deblurring.
\newblock In \emph{Proceedings of the IEEE/CVF Conference on Computer Vision and Pattern Recognition}, pages 149--165, 2021.

\bibitem[OpenAI(2023)]{openai2023gpt4}
OpenAI.
\newblock Gpt-4 technical report, 2023.

\bibitem[{\"O}zdenizci and Legenstein(2023)]{weatherdiff}
Ozan {\"O}zdenizci and Robert Legenstein.
\newblock Restoring vision in adverse weather conditions with patch-based denoising diffusion models.
\newblock \emph{IEEE Transactions on Pattern Analysis and Machine Intelligence}, 45\penalty0 (8):\penalty0 10346--10357, 2023.

\bibitem[Pan et~al.(2016)Pan, Sun, Pfister, and Yang]{deblur3}
Jinshan Pan, Deqing Sun, Hanspeter Pfister, and Ming-Hsuan Yang.
\newblock Blind image deblurring using dark channel prior.
\newblock In \emph{Proceedings of the IEEE conference on computer vision and pattern recognition}, pages 1628--1636, 2016.

\bibitem[Park et~al.(2019)Park, Yu, and Jeong]{denoise5}
Bumjun Park, Songhyun Yu, and Jechang Jeong.
\newblock Densely connected hierarchical network for image denoising.
\newblock In \emph{Proceedings of the IEEE/CVF conference on computer vision and pattern recognition workshops}, pages 0--0, 2019.

\bibitem[Potlapalli et~al.(2023)Potlapalli, Zamir, Khan, and Khan]{potlapalli2023promptir}
Vaishnav Potlapalli, Syed~Waqas Zamir, Salman Khan, and Fahad~Shahbaz Khan.
\newblock Promptir: Prompting for all-in-one blind image restoration.
\newblock \emph{arXiv preprint arXiv:2306.13090}, 2023.

\bibitem[Qian et~al.(2018)Qian, Tan, Yang, Su, and Liu]{raindrop}
Rui Qian, Robby~T Tan, Wenhan Yang, Jiajun Su, and Jiaying Liu.
\newblock Attentive generative adversarial network for raindrop removal from a single image.
\newblock In \emph{Proceedings of the IEEE conference on computer vision and pattern recognition}, pages 2482--2491, 2018.

\bibitem[Qin et~al.(2020)Qin, Wang, Bai, Xie, and Jia]{reside-6k}
Xu Qin, Zhilin Wang, Yuanchao Bai, Xiaodong Xie, and Huizhu Jia.
\newblock Ffa-net: Feature fusion attention network for single image dehazing.
\newblock In \emph{Proceedings of the AAAI conference on artificial intelligence}, pages 11908--11915, 2020.

\bibitem[Rasley et~al.(2020)Rasley, Rajbhandari, Ruwase, and He]{rasley2020deepspeed}
Jeff Rasley, Samyam Rajbhandari, Olatunji Ruwase, and Yuxiong He.
\newblock Deepspeed: System optimizations enable training deep learning models with over 100 billion parameters.
\newblock In \emph{Proceedings of the 26th ACM SIGKDD International Conference on Knowledge Discovery \& Data Mining}, pages 3505--3506, 2020.

\bibitem[Ren et~al.(2025)Ren, Li, Li, Wang, Guo, Zhao, Zhang, and Chen]{ren2025moe}
Yulin Ren, Xin Li, Bingchen Li, Xingrui Wang, Mengxi Guo, Shijie Zhao, Li Zhang, and Zhibo Chen.
\newblock Moe-diffir: Task-customized diffusion priors for universal compressed image restoration.
\newblock In \emph{European Conference on Computer Vision}, pages 116--134. Springer, 2025.

\bibitem[Roziere et~al.(2023)Roziere, Gehring, Gloeckle, Sootla, Gat, Tan, Adi, Liu, Remez, Rapin, et~al.]{roziere2023codellama}
Baptiste Roziere, Jonas Gehring, Fabian Gloeckle, Sten Sootla, Itai Gat, Xiaoqing~Ellen Tan, Yossi Adi, Jingyu Liu, Tal Remez, J{\'e}r{\'e}my Rapin, et~al.
\newblock Code llama: Open foundation models for code.
\newblock \emph{arXiv preprint arXiv:2308.12950}, 2023.

\bibitem[Russell and Norvig(2016)]{russell2016artificial}
Stuart~J Russell and Peter Norvig.
\newblock \emph{Artificial intelligence: a modern approach}.
\newblock Pearson, 2016.

\bibitem[Schick et~al.(2024)Schick, Dwivedi-Yu, Dess{\`\i}, Raileanu, Lomeli, Hambro, Zettlemoyer, Cancedda, and Scialom]{schick2024toolformer}
Timo Schick, Jane Dwivedi-Yu, Roberto Dess{\`\i}, Roberta Raileanu, Maria Lomeli, Eric Hambro, Luke Zettlemoyer, Nicola Cancedda, and Thomas Scialom.
\newblock Toolformer: Language models can teach themselves to use tools.
\newblock \emph{Advances in Neural Information Processing Systems}, 36, 2024.

\bibitem[Snell et~al.(2024)Snell, Lee, Xu, and Kumar]{snell2024scaling}
Charlie Snell, Jaehoon Lee, Kelvin Xu, and Aviral Kumar.
\newblock Scaling llm test-time compute optimally can be more effective than scaling model parameters.
\newblock \emph{arXiv preprint arXiv:2408.03314}, 2024.

\bibitem[Song et~al.(2020)Song, Meng, and Ermon]{ddim}
Jiaming Song, Chenlin Meng, and Stefano Ermon.
\newblock Denoising diffusion implicit models.
\newblock \emph{arXiv preprint arXiv:2010.02502}, 2020.

\bibitem[Sullivan et~al.(2012{\natexlab{a}})Sullivan, Ohm, Han, and Wiegand]{HM}
Gary~J Sullivan, Jens-Rainer Ohm, Woo-Jin Han, and Thomas Wiegand.
\newblock Overview of the high efficiency video coding (hevc) standard.
\newblock \emph{IEEE Transactions on circuits and systems for video technology}, 22\penalty0 (12):\penalty0 1649--1668, 2012{\natexlab{a}}.

\bibitem[Sullivan et~al.(2012{\natexlab{b}})Sullivan, Ohm, Han, and Wiegand]{sullivan2012overviewHEVC}
Gary~J Sullivan, Jens-Rainer Ohm, Woo-Jin Han, and Thomas Wiegand.
\newblock Overview of the high efficiency video coding (hevc) standard.
\newblock \emph{IEEE Transactions on circuits and systems for video technology}, 22\penalty0 (12):\penalty0 1649--1668, 2012{\natexlab{b}}.

\bibitem[Sun et~al.(2024)Sun, Li, Liu, Chen, Pei, Zou, Yan, and Yang]{sun2024coser}
Haoze Sun, Wenbo Li, Jianzhuang Liu, Haoyu Chen, Renjing Pei, Xueyi Zou, Youliang Yan, and Yujiu Yang.
\newblock Coser: Bridging image and language for cognitive super-resolution.
\newblock In \emph{Proceedings of the IEEE/CVF Conference on Computer Vision and Pattern Recognition}, pages 25868--25878, 2024.

\bibitem[Tao et~al.(2018)Tao, Gao, Shen, Wang, and Jia]{deblur2}
Xin Tao, Hongyun Gao, Xiaoyong Shen, Jue Wang, and Jiaya Jia.
\newblock Scale-recurrent network for deep image deblurring.
\newblock In \emph{Proceedings of the IEEE conference on computer vision and pattern recognition}, pages 8174--8182, 2018.

\bibitem[Timofte et~al.(2017)Timofte, Agustsson, Van~Gool, Yang, and Zhang]{Flickr2K}
Radu Timofte, Eirikur Agustsson, Luc Van~Gool, Ming-Hsuan Yang, and Lei Zhang.
\newblock Ntire 2017 challenge on single image super-resolution: Methods and results.
\newblock In \emph{Proceedings of the IEEE conference on computer vision and pattern recognition workshops}, pages 114--125, 2017.

\bibitem[Touvron et~al.(2023)Touvron, Martin, Stone, Albert, Almahairi, Babaei, Bashlykov, Batra, Bhargava, Bhosale, et~al.]{touvron2023llama2}
Hugo Touvron, Louis Martin, Kevin Stone, Peter Albert, Amjad Almahairi, Yasmine Babaei, Nikolay Bashlykov, Soumya Batra, Prajjwal Bhargava, Shruti Bhosale, et~al.
\newblock Llama 2: Open foundation and fine-tuned chat models.
\newblock \emph{arXiv preprint arXiv:2307.09288}, 2023.

\bibitem[Vaswani(2017)]{vaswani2017attention}
A Vaswani.
\newblock Attention is all you need.
\newblock \emph{Advances in Neural Information Processing Systems}, 2017.

\bibitem[Wang et~al.(2018)Wang, Wei, Yang, and Liu]{wang2018gladnetlow-light}
Wenjing Wang, Chen Wei, Wenhan Yang, and Jiaying Liu.
\newblock Gladnet: Low-light enhancement network with global awareness.
\newblock In \emph{2018 13th IEEE international conference on automatic face \& gesture recognition (FG 2018)}, pages 751--755. IEEE, 2018.

\bibitem[Wang et~al.(2021)Wang, Xie, Dong, and Shan]{realesrgan}
Xintao Wang, Liangbin Xie, Chao Dong, and Ying Shan.
\newblock Real-esrgan: Training real-world blind super-resolution with pure synthetic data.
\newblock In \emph{Proceedings of the IEEE/CVF international conference on computer vision}, pages 1905--1914, 2021.

\bibitem[Wang et~al.(2022)Wang, Cun, Bao, Zhou, Liu, and Li]{wang2022uformer}
Zhendong Wang, Xiaodong Cun, Jianmin Bao, Wengang Zhou, Jianzhuang Liu, and Houqiang Li.
\newblock Uformer: A general u-shaped transformer for image restoration.
\newblock In \emph{Proceedings of the IEEE/CVF conference on computer vision and pattern recognition}, pages 17683--17693, 2022.

\bibitem[Wei et~al.()Wei, Wang, Yang, and Liu]{lol}
C Wei, W Wang, W Yang, and J Liu.
\newblock Deep retinex decomposition for low-light enhancement. arxiv 2018.
\newblock \emph{arXiv preprint arXiv:1808.04560}.

\bibitem[Wei et~al.(2023)Wei, Zhang, Ren, Xu, Hong, Yang, Yan, and Wang]{claritychatgpt}
Yanyan Wei, Zhao Zhang, Jiahuan Ren, Xiaogang Xu, Richang Hong, Yi Yang, Shuicheng Yan, and Meng Wang.
\newblock Clarity chatgpt: An interactive and adaptive processing system for image restoration and enhancement.
\newblock \emph{arXiv preprint arXiv:2311.11695}, 2023.

\bibitem[Wieschollek et~al.(2017)Wieschollek, Hirsch, Scholkopf, and Lensch]{deblur4}
Patrick Wieschollek, Michael Hirsch, Bernhard Scholkopf, and Hendrik Lensch.
\newblock Learning blind motion deblurring.
\newblock In \emph{Proceedings of the IEEE international conference on computer vision}, pages 231--240, 2017.

\bibitem[Wu et~al.(2024)Wu, Zhu, Zhang, Zhang, Chen, Liao, Li, Wang, Sun, Yan, Liu, Zhai, Wang, and Lin]{coinstruct}
Haoning Wu, Hanwei Zhu, Zicheng Zhang, Erli Zhang, Chaofeng Chen, Liang Liao, Chunyi Li, Annan Wang, Wenxiu Sun, Qiong Yan, Xiaohong Liu, Guangtao Zhai, Shiqi Wang, and Weisi Lin.
\newblock Towards open-ended visual quality comparison, 2024.

\bibitem[Xia et~al.(2023)Xia, Zhang, Wang, Wang, Wu, Tian, Yang, and Van~Gool]{xia2023diffir}
Bin Xia, Yulun Zhang, Shiyin Wang, Yitong Wang, Xinglong Wu, Yapeng Tian, Wenming Yang, and Luc Van~Gool.
\newblock Diffir: Efficient diffusion model for image restoration.
\newblock In \emph{Proceedings of the IEEE/CVF International Conference on Computer Vision}, pages 13095--13105, 2023.

\bibitem[Xing et~al.(2020)Xing, Xu, Li, and Guan]{dejpeg4}
Qunliang Xing, Mai Xu, Tianyi Li, and Zhenyu Guan.
\newblock Early exit or not: Resource-efficient blind quality enhancement for compressed images.
\newblock In \emph{European Conference on Computer Vision}, pages 275--292. Springer, 2020.

\bibitem[Yang et~al.(2023)Yang, Dong, Liu, Li, Wang, Jiang, Tan, Kang, Zhang, Zhou, et~al.]{embodiedoctopus}
Jingkang Yang, Yuhao Dong, Shuai Liu, Bo Li, Ziyue Wang, Chencheng Jiang, Haoran Tan, Jiamu Kang, Yuanhan Zhang, Kaiyang Zhou, et~al.
\newblock Octopus: Embodied vision-language programmer from environmental feedback.
\newblock \emph{arXiv preprint arXiv:2310.08588}, 2023.

\bibitem[Yang et~al.(2017)Yang, Tan, Feng, Liu, Guo, and Yan]{rain100H}
Wenhan Yang, Robby~T Tan, Jiashi Feng, Jiaying Liu, Zongming Guo, and Shuicheng Yan.
\newblock Deep joint rain detection and removal from a single image.
\newblock In \emph{Proceedings of the IEEE conference on computer vision and pattern recognition}, pages 1357--1366, 2017.

\bibitem[Ye et~al.(2023)Ye, Xu, Ye, Yan, Liu, Qian, Zhang, Huang, and Zhou]{ye2023mplug}
Qinghao Ye, Haiyang Xu, Jiabo Ye, Ming Yan, Haowei Liu, Qi Qian, Ji Zhang, Fei Huang, and Jingren Zhou.
\newblock mplug-owl2: Revolutionizing multi-modal large language model with modality collaboration.
\newblock \emph{arXiv preprint arXiv:2311.04257}, 2023.

\bibitem[Yu et~al.(2024)Yu, Gu, Li, Hu, Kong, Wang, He, Qiao, and Dong]{yu2024SUPIR}
Fanghua Yu, Jinjin Gu, Zheyuan Li, Jinfan Hu, Xiangtao Kong, Xintao Wang, Jingwen He, Yu Qiao, and Chao Dong.
\newblock Scaling up to excellence: Practicing model scaling for photo-realistic image restoration in the wild.
\newblock \emph{arXiv preprint arXiv:2401.13627}, 2024.

\bibitem[Yu et~al.(2018)Yu, Dong, Lin, and Loy]{rl-restore}
Ke Yu, Chao Dong, Liang Lin, and Chen~Change Loy.
\newblock Crafting a toolchain for image restoration by deep reinforcement learning.
\newblock In \emph{Proceedings of the IEEE conference on computer vision and pattern recognition}, pages 2443--2452, 2018.

\bibitem[Zamir et~al.(2022)Zamir, Arora, Khan, Hayat, Khan, and Yang]{zamir2022restormer}
Syed~Waqas Zamir, Aditya Arora, Salman Khan, Munawar Hayat, Fahad~Shahbaz Khan, and Ming-Hsuan Yang.
\newblock Restormer: Efficient transformer for high-resolution image restoration.
\newblock In \emph{Proceedings of the IEEE/CVF conference on computer vision and pattern recognition}, pages 5728--5739, 2022.

\bibitem[Zhang et~al.(2017)Zhang, Zuo, Chen, Meng, and Zhang]{denoise2}
Kai Zhang, Wangmeng Zuo, Yunjin Chen, Deyu Meng, and Lei Zhang.
\newblock Beyond a gaussian denoiser: Residual learning of deep cnn for image denoising.
\newblock \emph{IEEE transactions on image processing}, 26\penalty0 (7):\penalty0 3142--3155, 2017.

\bibitem[Zhang et~al.(2018{\natexlab{a}})Zhang, Zuo, and Zhang]{denoise4}
Kai Zhang, Wangmeng Zuo, and Lei Zhang.
\newblock Ffdnet: Toward a fast and flexible solution for cnn-based image denoising.
\newblock \emph{IEEE Transactions on Image Processing}, 27\penalty0 (9):\penalty0 4608--4622, 2018{\natexlab{a}}.

\bibitem[Zhang et~al.(2021)Zhang, Liang, Van~Gool, and Timofte]{bsrgan}
Kai Zhang, Jingyun Liang, Luc Van~Gool, and Radu Timofte.
\newblock Designing a practical degradation model for deep blind image super-resolution.
\newblock In \emph{Proceedings of the IEEE/CVF International Conference on Computer Vision}, pages 4791--4800, 2021.

\bibitem[Zhang et~al.(2022{\natexlab{a}})Zhang, Li, Liang, Cao, Zhang, Tang, Timofte, and Van~Gool]{denoise3}
Kai Zhang, Yawei Li, Jingyun Liang, Jiezhang Cao, Yulun Zhang, Hao Tang, Radu Timofte, and Luc Van~Gool.
\newblock Practical blind denoising via swin-conv-unet and data synthesis.
\newblock \emph{arXiv e-prints}, pages arXiv--2203, 2022{\natexlab{a}}.

\bibitem[Zhang et~al.(2022{\natexlab{b}})Zhang, Ren, Luo, Lai, Stenger, Yang, and Li]{deblur5}
Kaihao Zhang, Wenqi Ren, Wenhan Luo, Wei-Sheng Lai, Bj{\"o}rn Stenger, Ming-Hsuan Yang, and Hongdong Li.
\newblock Deep image deblurring: A survey.
\newblock \emph{International Journal of Computer Vision}, 130\penalty0 (9):\penalty0 2103--2130, 2022{\natexlab{b}}.

\bibitem[Zhang et~al.(2011)Zhang, Wu, Buades, and Li]{mcmaster}
Lei Zhang, Xiaolin Wu, Antoni Buades, and Xin Li.
\newblock Color demosaicking by local directional interpolation and nonlocal adaptive thresholding.
\newblock \emph{Journal of Electronic imaging}, 20\penalty0 (2):\penalty0 023016--023016, 2011.

\bibitem[Zhang et~al.(2018{\natexlab{b}})Zhang, Tian, Kong, Zhong, and Fu]{rdn}
Yulun Zhang, Yapeng Tian, Yu Kong, Bineng Zhong, and Yun Fu.
\newblock Residual dense network for image super-resolution.
\newblock In \emph{Proceedings of the IEEE conference on computer vision and pattern recognition}, pages 2472--2481, 2018{\natexlab{b}}.

\bibitem[Zhu et~al.(2024)Zhu, Gu, You, Qiao, and Dong]{agenticIR}
Kaiwen Zhu, Jinjin Gu, Zhiyuan You, Yu Qiao, and Chao Dong.
\newblock An intelligent agentic system for complex image restoration problems.
\newblock \emph{arXiv preprint arXiv:2410.17809}, 2024.

\end{thebibliography}
}
\clearpage

\maketitlesupplementary

\section{More Details about Restoration Tools}
\label{sec:ab1}
In this section, we provide more details to support our claims in Section 3.2 of the main paper.

\begin{figure*}[!ht]
    \centering
    \includegraphics[width=1\textwidth]{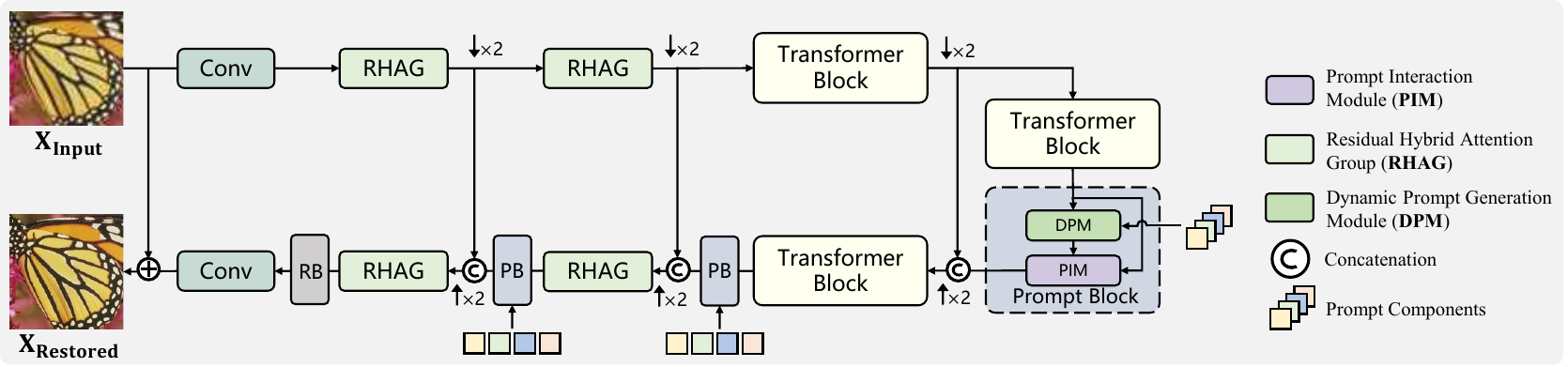}
    \caption{The illustration of network architecture of restoration model. We adopt the enhanced version of PromptIR~\cite{potlapalli2023promptir} from~\cite{li2024promptcir}. ``RB'' denotes for refinement block, ``PB'' denotes for prompt block, respectively.}
    \label{fig:ab_arch}
    % \FloatBarrier
\end{figure*}

\subsection{Network Architecture of Restoration Model}

We propose a three-stage training recipe for the construction of restoration tools. In the first stage, we aim to learn a general-purpose restoration model that encompasses common knowledge across various tasks. Therefore, we follow promising prompt learning-based works~\cite{potlapalli2023promptir} and adopt an enhanced version~\cite{li2024promptcir} to further improve the representative abilities of the restoration model. The detailed model architecture is demonstrated in Figure~\ref{fig:ab_arch}.

\subsection{LoRA Fine-tuning of Restoration Tools}

\begin{figure*}[]
    \centering
    \includegraphics[width=1\linewidth]{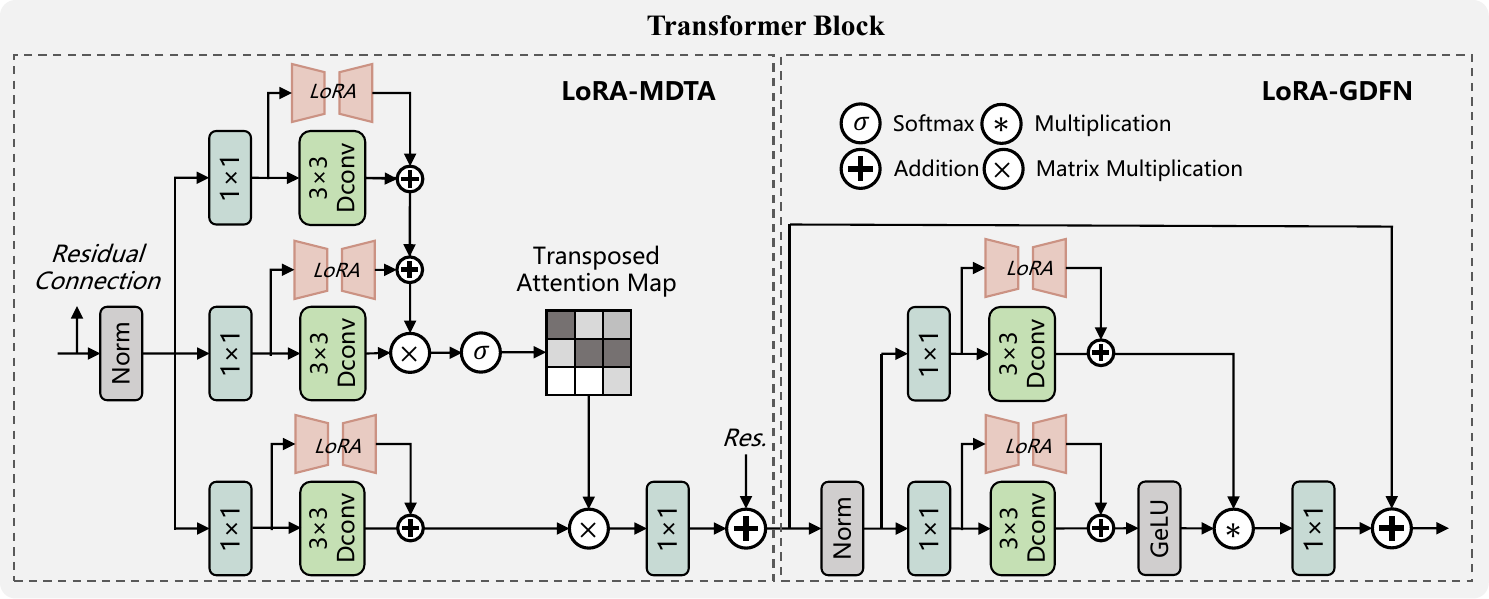}
    \caption{The implementation details of LoRA fine-tuning. We take the transformer block~\cite{potlapalli2023promptir} here for an example. As multi-DConv head transposed self-attention (MDTA) leverages depth-wise convolution to generate Q, K, V matrices, we add LoRA on these layers. For residual hybrid attention group (RHAG), we add LoRA on linear layers that related to Q, K, V matrices generation. Additionally, following common implementation, we add LoRA layers in the feedforward module.}
    \label{fig:ab_lora}
\end{figure*}

We employ LoRA~\cite{hu2021lora} to effectively build task-specific restoration tools based on the well-trained model from Stage I. We follow the general implementation and add low-rank matrices to the attention block layers that generate Q, K, and V, respectively. Moreover, the LoRA is added to the Linear layer in the feedforward block. The similar implementation is adopted for RHAG~\cite{HAT}. The details are demonstrated in Figure~\ref{fig:ab_lora}. 

\subsection{The Generation of Hybrid Distortion}
\label{sec:pipeline}

\begin{figure*}[]
    \centering
    \includegraphics[width=1\linewidth]{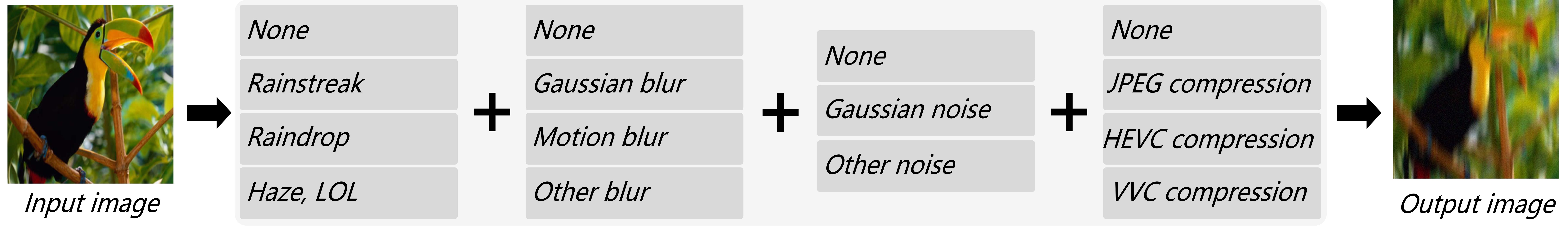}
    \caption{The distortion pipeline we used to synthesize hybrid degradations. Notably, for rainstreak, raindrop, haze, and low-light images, we only add Gaussian noise or JPEG distortions. ``Other blur'' and ``other noise'' refer to additional types of blur and noise implemented in Real-ESRGAN~\cite{realesrgan}. During synthesis, we ensure that each image contains at least two types of distortions.}
    \label{fig:ab_pipeline}
\end{figure*}

To tailor the hybrid restoration tool, we further fine-tune the restoration model on online-generated mix-degraded image pairs. As the degradation pipeline in Real-ESRGAN~\cite{realesrgan} and BSRGAN~\cite{bsrgan} shows promising results on hybrid and real-world distortion removal, we follow such design and build a degradation pipeline that includes our 10 distortions in Figure~\ref{fig:ab_pipeline}. Considering that rainstreak, raindrop, haze and low light are intrinsically real-world distortions, we only add noise/JPEG on these four types of distortions to form the hybrid distortion. Among the 10 distortions, we use existing datasets for rainstreak, raindrop, haze, and low-light. For the other six distortions, we synthesize image pairs. The specific synthesis methods are as follows:
\begin{itemize}
    \item Gaussian blur: We add Gaussian blur with sigma value ranging from 0.2 to 4.
    \item Motion blur: We follow the implementation in~\cite{duan2025uniprocessor}.
    \item Gaussian noise: We add Gaussian noise with intensity value ranging from 15 to 50.
    \item JPEG compression: We add JPEG compression with quality factor between 10 and 40.
    \item HEVC compression: We add HEVC compression (HM-18.0) with three quality factor 32, 37, and 42, where a higher value demonstrates a worse image quality.
    \item VVC compression: Similar as HEVC, we add VVC vompression (VTM-21.0) with three quality factor 32, 37, and 42.
\end{itemize}

\subsection{The Construction of Instruction Tuning Dataset}

In this section, we provide more details about the construction of our instruction tuning dataset for SlowAgent and FeedbackAgent, respectively. Furthermore, we provide examples for a more intuitive illustration of the constructed instruction tuning dataset.

\subsubsection{Instruction Tuning Dataset for SlowAgent}

Once the FastAgent categories the user prompt as the ambiguous one, it will invoke the SlowAgent to automatically finish the restoration process. Consequently, the task of the SlowAgent is to detect distortions and select the appropriate restoration tool. Therefore, we build the instruction tuning dataset based on the distortion type of the image and the according restoration tool. As described in Sections 3.3 and 4.1 of the main paper, we construct the instruction-tuning dataset based on 11 types of distortions (\ie, 10 single distortion types and one hybrid distortion type). For each image-text pairs, we use the following format: ``[User: \textless Question\textgreater \textless image\textgreater.], [Assistant: DISTORTION: \textless type\textgreater. CALL: de-\textless type\textgreater tool.]'', where ``Question'' is randomly chosen from Table~\ref{tab:question}, and ``type'' denotes for distortion type. Notably, to prevent the multimodal large language model (MLLM) from overfitting to the questions rather than the image distortion identification, we used GPT to randomly generate 20 different questions.

\begin{table}[!h]
\centering
\caption{List of Questions we utilized to construct the instruction tuning dataset for SlowAgent.}
\resizebox{\linewidth}{!}{
\begin{tabular}{l}
\toprule
\textbf{Questions} \\ 
\midrule
What is the distortion type of this image? \\
What kind of distortion is present in this image? \\
What type of image distortion can be observed here? \\
What distortion effect is visible in this image? \\
Can you identify the distortion in this image? \\
What is the nature of the distortion in this image? \\
What type of distortion has affected this image? \\
What form of distortion is evident in this image? \\
How is this image distorted? \\
What kind of image distortion does this show? \\
What kind of visual distortion is in this image? \\
What distortion does this image exhibit? \\
What is the specific distortion type in this image? \\
How is this image distorted visually? \\
What kind of alteration or distortion appears in this image? \\
What type of distortion can be seen in this image? \\
What image distortion effect is noticeable here? \\
What is the distortion pattern in this image? \\
Can you describe the distortion present in this image? \\
What distortion characteristic is evident in this image? \\
\bottomrule
\end{tabular}}
\label{tab:question}
\end{table}

Additionally, to enable the SlowAgent to robustly identify distortion types in images, we constructed a dataset containing 70k samples. We generate 5k image-text pairs for each single distortion and 20k pairs for hybrid distortions. For the hybrid distortions, we include 2k image-text pairs for rainstreak, raindrop, haze, and low-light with noise/JPEG, respectively. We use distortion pipeline described in Section~\ref{sec:pipeline} to generate another 12k hybrid distortions image-text pairs.

\subsubsection{Instruction Tuning Dataset for FeedbackAgent}

In our HybridAgent system, the SlowAgent is responsible for recognizing distortions in images. However, it lacks the ability to determine whether the restoration process should be terminated or continued. Therefore, we develop a FeedbackAgent to determine whether an image is clean, providing support for the restoration process. For each image-text pairs, we use the following format: ``[User: This is a restored image.\textless image\textgreater RESTORATION HISTORY: de-\textless type\textgreater. Is it clean now?], [Assistant: Yes. CALL: END.] or [Assistant: No. CALL: SlowAgent.]'', where ``END'' indicates that the restoration process is finished. Follow the description in Section 3.3 of the main paper, we construct 30k image-text pairs for ``clean'' and 33k image-text pairs for ``not clean''. For ``clean'' samples, we generate 25k image-text pairs for (i) single-distorted images restored using the correct tool (2.5k pairs for each single distortion), 2.5k pairs for (ii) hybrid-distorted images restored with the hybrid restoration tool, and 2.5k pairs for (iii) single-distorted images restored by the hybrid restoration tool. For ``not clean'' samples, we generate 8k image-text pairs for (i) single-distorted images restored with an incorrect tool, and 25k image-text pairs for (ii) hybrid-distorted images restored with a single restoration tool (2.5k pairs for each task-specific restoration tool).

\section{More Implementation Details}
\label{sec:ab2}

\subsection{Instruction Tuning for HybridAgent}
\label{sec:instructiontuning}

Current studies on MLLMs generally focus on building models that excel in diverse tasks~\cite{bai2023qwen,chen2023internvl,openai2023gpt4,llava}. However, this generalization often limits their performance on specialized tasks requiring expert knowledge, such as IR. To adapt MLLMs for the roles of SlowAgent and FeedbackAgent, we follow previous works~\cite{llava,ye2023mplug,coinstruct} and adopt instruction tuning.

\noindent\textbf{SlowAgent.} Since the primary task of the SlowAgent is to detect distortions and select the appropriate restoration tool, the instruction-tuning dataset must encompass a broad range of distortion types, with corresponding text outputs designed as invocation commands for restoration tools. Specifically, to cover the majority of real application scenarios, we construct our dataset with 10 distortions: noise, gaussian blur, motion blur, JPEG, HEVC~\cite{HM}, VVC~\cite{VTM}, rainstreak, raindrop, haze, low light. Additionally, the combination of these distortions is considered the 11th type. We provide details about the combination in the \textbf{supplementary.} Since SlowAgent's distortion recognition should be independent of image content and resolution, we apply random rotations and flips as augmentation. We randomly crop the image to a resolution ranging between $224\times224$ and $784\times784$. Our instruction tuning dataset for the SlowAgent includes 70k image-text pairs, enabling the tuned model to robustly identify image distortion types. More details are given in the \textbf{supplementary}.

\noindent\textbf{FeedbackAgent.} On the other hand, since no existing MLLM can reliably assess whether a restored image is clean, instruction tuning is essential for the FeedbackAgent. To address this, we constructed an additional dataset with approximately 60k image-text pairs, where images are labeled as ``clean'' or ``not clean''. We define ``clean'' as: (i) single-distorted images restored using the correct tool, (ii) hybrid-distorted images restored with the hybrid restoration tool, or (iii) single-distorted images restored by the hybrid restoration tool. Conversely, ``not clean'' includes: (i) single-distorted images restored with an incorrect tool, or (ii) hybrid-distorted images restored with a single restoration tool.

\subsection{Training Details}

The training of HybridAgent mainly contains two parts: (i) The construction of restoration tools, and (ii) The instruction tuning of SlowAgent and FeedbackAgent. To build restoration tools, we first train a prompt learning-based all-in-one model~\cite{li2024promptcir} across 10 distortions, with an initial learning rate of 2e-4 and gradually decay to 1e-6 with cosine annealing~\cite{loshchilov2016sgdr}. AdamW optimizer is adopted to train the model for 600k iterations. During training, the image pairs are cropped into $128\times 128$ patches with random horizontal and vertical flips as the data augmentation. Subsequently, we leverage LoRA with a rank of 8 to tailor our task-specific restoration tools. For each tool, the model with LoRA weights is further optimized for 100k iterations with a fixed learning rate of 5e-5. Consequently, to build a hybrid restoration tool, we further fine-tune a new set of LoRA weights with mixed degradations for 200k iterations, where a fixed learning rate of 1e-4 is adopted. For Stage I, we utilized 8 RTX 3090 GPUs for training, with a total batch size of 32. For Stage II and III, we utilized 4 4090D with a total batch size of 16. We provide the training times in Table~\ref{tab:traintime}. Notably, we generate the mixed-degraded samples online in Stage III, where the compression processes of HEVC~\cite{sullivan2012overviewHEVC} and VVC~\cite{bross2021overviewVVC} consume a significant amount of time.

\begin{table}[]
\centering
\caption{Training consumptions of the restoration tools. Notably, we generate the mixed-degraded samples online in Stage III for storage saving purpose, where the compression processes of HEVC and VVC consume a significant amount of time.}
\begin{tabular}{l|cc}
\toprule
 & Trainable Params. (M) & Training Hours \\ \midrule
Stage I & 34.79 & 87.71 \\
Stage II & 6.49 & 8.98 \\
Stage III & 6.36 & 31.47 \\ \bottomrule
\end{tabular}
\label{tab:traintime}
\end{table}

We instruct tuning Co-Instruct~\cite{coinstruct} model for our HybridAgent, as it excels at distortion-related question-answering tasks. For both agents, we fine-tune the MLLM using 4 RTX 4090D GPUs with a total batch size of 256 using LoRA~\cite{hu2021lora}. The initial learning rate is set to 1e-4 and gradually decayed to 0 using cosine annealing. Each agent is trained for 2 epochs, with SlowAgent taking approximately 2.5 hours and FeedbackAgent about 1.5 hours. Deepspeed~\cite{rasley2020deepspeed} is utilized to accelerate the training process.

\subsection{More Details about Testset}

Beyond single distortion removal, it is crucial to evaluate the hybrid distortion removal capabilities of HybridAgent. To this end, we generate a total of 200 mix-degraded images, with details provided in Table~\ref{tab:hybridnumber}. For the first six rows, we select 20 images from the combined datasets of CBSD68~\cite{cbsd}, Urban100~\cite{urban100}, Kodak24~\cite{kodak}, and McMaster~\cite{mcmaster} as ground truth. For the remaining rows, we select 10 distorted images from Rain100H~\cite{rain100H}, RainDrop~\cite{raindrop}, RESIDE-6k~\cite{reside-6k}, and LOL~\cite{lol}, and further add noise or JPEG artifacts.

\begin{table}[]
    \centering
    \caption{Number of samples for the hybrid distortion testset. In total we generate 200 images for the evaluation of hybrid distortion removal.}
    \begin{tabular}{l r}
\toprule
\textbf{Hybrid Distortion} & \textbf{Number of Samples} \\
\midrule
Blur + JPEG                & 20 \\
Blur + Noise               & 20 \\
Blur + Noise + JPEG        & 20 \\
Motionblur + JPEG          & 20 \\
Motionblur + Noise         & 20 \\
Motionblur + Noise + JPEG  & 20 \\
Rainstreak + JPEG          & 10 \\
Rainstreak + Noise         & 10 \\
Raindrop + JPEG            & 10 \\
Raindrop + Noise           & 10 \\
Haze + JPEG                & 10 \\
Haze + Noise               & 10 \\
Low light + JPEG           & 10 \\
Low light + Noise          & 10 \\
\bottomrule
\end{tabular}
\label{tab:hybridnumber}
\end{table}

\subsection{More details about User Prompt}

Users can provide various textual prompts to HybridAgent. To synthesize such prompts and evaluate the distinguishing capabilities of FastAgent, we use GPT-4 to generate a total of 220 diverse user prompts. We provide some samples in Table~\ref{tab:textprompt}. \textit{Notably, we assume the user has precise knowledge of the distortion type to carry out direct prompts.}

\begin{table}[!t]
\centering
\caption{Examples of textual user prompts generated by GPT-4. Notably, we assume the user has precise knowledge of the distortion type to carry out direct prompts.}
\begin{tabular}{p{0.2\linewidth} p{0.7\linewidth}}
\toprule
\textbf{Distortion} & \textbf{Textual Prompts} \\
\midrule
\multirow{4}{*}{Noise}                & Please remove the grain from this image. \\
              & The speckles in this photo need to be cleared up. \\
       & Please fix the random spots in this image. \\ \midrule
\multirow{6}{*}{HEVC}          & Can you reduce the H.265 artifacts to improve the picture’s clarity? \\
         & The HM compression makes the image look rough; can you fix it? \\
  & Can you remove the HEVC artifacts for a clearer image? \\ \midrule
\multirow{6}{*}{Haze}          & Please reduce the haze that blurs the scene. \\
         & I’d prefer the photo to be haze-free for better contrast. \\
           & I’d like the image to look vibrant, free from the dull haze. \\ \midrule
\multirow{4}{*}{Ambiguous}            & Please fix this image. \\
               & This image does not look good, please help me. \\
              & Can you help me enhance this image? \\
\bottomrule
\end{tabular}
\label{tab:textprompt}
\end{table}

\begin{table}[]
% \vspace{-3mm}
\centering
\caption{Comparisons of performance on hybrid removal of mixed distortions. We evaluate the PSNR$\uparrow$/SSIM$\uparrow$/LPIPS$\downarrow$.}
\resizebox{\linewidth}{!}{
\begin{tabular}{l|ccc}
\toprule
 & Rainstreak + Noise & Rainstreak + JPEG & Noise + JPEG \\ \midrule
From Scratch & 26.31/0.760/0.288 & 25.35/0.755/0.266 & 30.60/0.879/0.102 \\
Stage III & 26.49/0.766/0.203 & 25.44/0.764/0.242 & 30.64/0.882/0.096  \\ \bottomrule
\end{tabular}}
\label{tab:stage3vsscratch}
\end{table}

\begin{table}[]
% \vspace{-3mm}
\centering
\small
\caption{Comparisons of performance on whether the parameters of prompt components are reinitialized in Stage II. We evaluate the PSNR$\uparrow$/SSIM$\uparrow$.}

\begin{tabular}{l|ccc}
\toprule
 & De-Noise & De-Raindrop & De-Low light \\ \midrule
(i) & 30.76/0.877 & 30.35/0.914 & 22.61/0.828 \\
(ii) & 30.72/0.876 & 29.12/0.907 & 22.37/0.827  \\
(iii) & 30.70/0.876 & 27.89/0.889 & 21.88/0.816  \\ \bottomrule
\end{tabular}
\label{tab:reinitialization}
\end{table}

\begin{table*}[]
\vspace{-3mm}
\centering
\caption{Comparisons of performance on single distortion removal between Stage I's base model and Stage II's task-specific restoration tools. We evaluate the PSNR$\uparrow$/SSIM$\uparrow$.}
\vspace{-1mm}
\resizebox{\textwidth}{!}{
\begin{tabular}{@{}l|cccccccccc@{}}
\toprule
 & De-noise & De-blur & De-motionblur & De-jpeg & De-HEVC & De-VVC & De-rainstreak & De-raindrop & De-haze & De-low light \\ \midrule
Stage I & 30.63/0.874 & 28.97/0.834 & 23.28/0.709 & 29.17/0.862 & 26.73/0.779 & 27.09/0.792 & 28.33/0.867  & 26.05/0.893 & 16.14/0.777 & 21.49/0.810 \\
Stage II & 30.76/0.877 & 30.65/0.853 & 23.80/0.720 & 30.21/0.876 & 27.58/0.785 & 27.69/0.794 & 30.05/0.894 & 30.35/0.914 & 29.93/0.960 & 22.61/0.828 \\ \bottomrule
\end{tabular}}
\label{tab:stage2vs1}
\vspace{-3mm}
\end{table*}

\section{Analysis of Three-Stage Training}
\label{sec:ab3}

\subsection{Effectiveness of Stage II}

\begin{figure}
% \vspace{-4mm}
    \centering
    \includegraphics[width=1\linewidth]{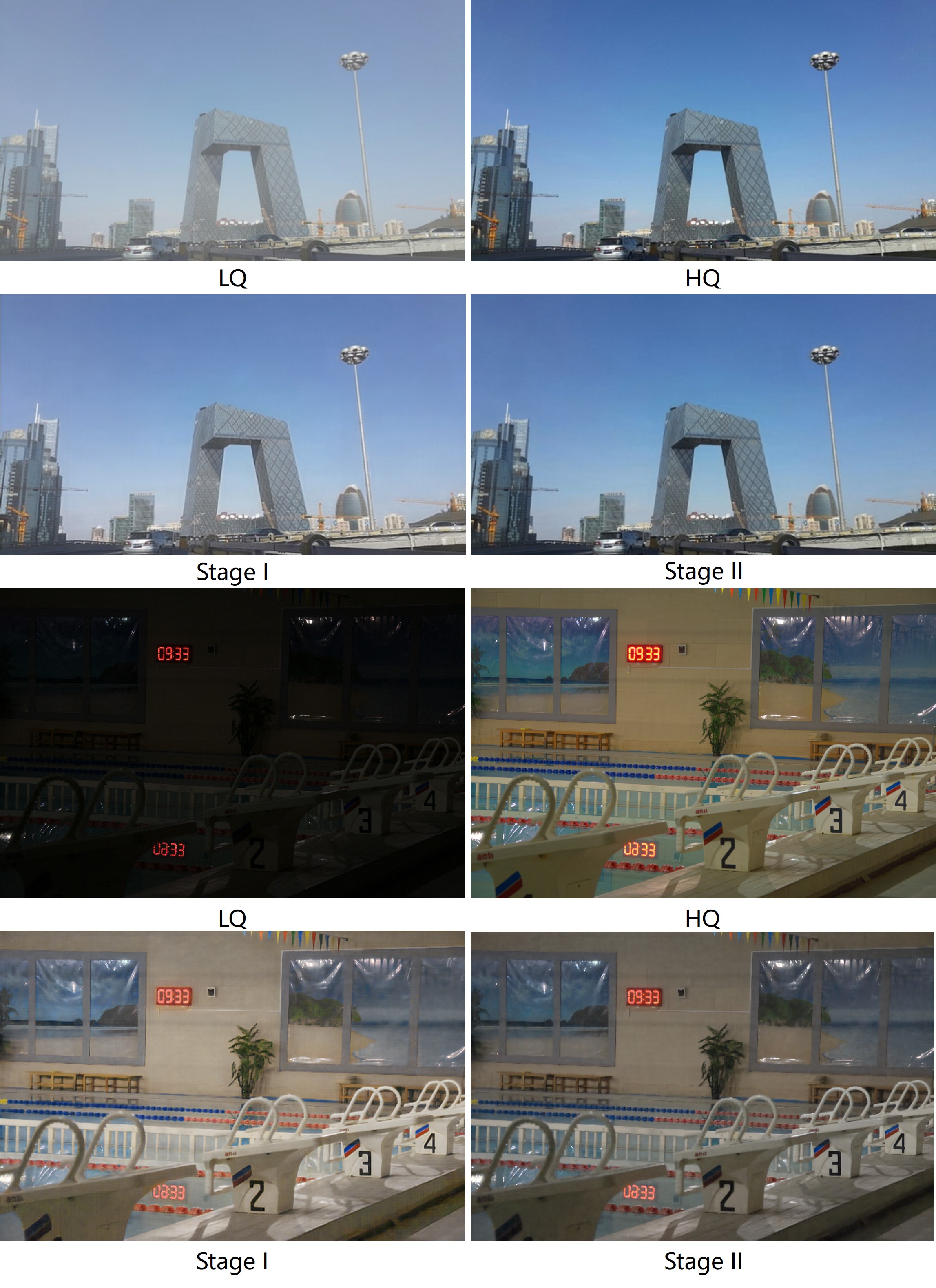}
    \caption{Qualitative comparisons between Stage I's and Stage II's restoration results on de-haze (top) and low light enhancement (bottom).}
    \label{fig:ab1}
    \vspace{-4mm}
\end{figure}

To construct restoration tools that not only share common knowledge across different distortion removal tasks but also possess task-specific expertise, we propose a three-stage training recipe in Section 3.2 of the main paper. To demonstrate the effectiveness of Stage II, we compare the performance between base model and task-specific model on 10 distortions. As shown in Table~\ref{tab:stage2vs1}, task-specific restoration tools achieve performance improvements over the base model. Notably, task-specific tools outperform the base model, especially for rain, haze, and low-light distortions. This further highlights the importance of task-specific tools in enabling HybridAgent to effectively handle various distortions. As demonstrated in Figure~\ref{fig:ab1}, although the base model learns common knowledge across various distortions, its ability to handle specific distortions may be compromised. For instance, in low-light enhancement, the image becomes overly bright. Therefore, it is crucial for HybridAgent to leverage task-specific restoration tools to effectively handle various distortions.

On the other hand, to highlight the importance of common knowledge shared across different tasks, we compare step-by-step removal of three hybrid distortions using Stage II's restoration tools with task-specific models trained from scratch. As demonstrated in Table~\ref{tab:stage2vsscratch}, Stage II's restoration tools achieve higher performance, indicating the effectiveness of task-common knowledge learning in Stage I. Additionally, we only have to storage light-weight LoRA weights for restoration tools, which is resource-friendly towards numerous of distortions.

\begin{table}[!h]
\vspace{-3mm}
\caption{Comparisons of performance on step-by-step removal of hybrid distortions. We evaluate the PSNR$\uparrow$/SSIM$\uparrow$/LPIPS$\downarrow$.}
\resizebox{\linewidth}{!}{
\begin{tabular}{l|ccc}
\toprule
 & Rainstreak + Noise & Rainstreak + JPEG & Noise + JPEG \\ \midrule
From Scratch & 23.13/0.617/0.372 & 22.04/0.685/0.331 & 30.19/0.813/0.185 \\
Stage II & 23.36/0.622/0.351 & 22.18/0.692/0.300 & 30.23/0.816/0.163  \\ \bottomrule
\end{tabular}}
\label{tab:stage2vsscratch}
\vspace{-3mm}
\end{table}

\subsection{Effectiveness of Stage III}

To demonstrate the advantage of Stage III training for the hybrid restoration tool, we compare its performance with models trained from scratch with the same mix-degradation pipeline. As shown in Table~\ref{tab:stage3vsscratch}, the hybrid restoration tool trained in Stage III outperforms the model trained from scratch, especially for complex distortions (\eg, rainstreak).

\subsection{Effectiveness of Parameter Reinitialization}

Prompt components serve as conditional information, enabling the restoration model to identify distortions and execute the appropriate removal process. Therefore, reinitializing the parameters of prompt components in Stage II is essential to prevent the model from generating inaccurate conditional information. To validate this, we conduct experiments on three settings: (i) As demonstrated in the main paper, we reinitialize the parameters of the prompt components; (ii) We initialize the prompt components with the respective parameters from Stage I; (iii) We initialize the prompt components with the respective parameters from Stage I and keep them fixed during LoRA fine-tuning. The results are demonstrated in Table~\ref{tab:reinitialization}. Comparing between (i) and (ii), we conclude that reinitialize prompt components are essential for the construction of task-specific restoration tools, especially for difficult distortions such as raindrop and low-light. \textit{Furthermore, we conclude that LoRA fine-tuning is efficient than full fine-tuning the model.} By comparing (ii) and (iii), we further validate that the prompt components from Stage I are not suitable for task-specific learning, as they generate inaccurate distortion guidance, leading to suboptimal results.

\section{Qualitative Results of All-in-One Methods}
\label{sec:ab4}

\begin{figure*}
    \centering
    \includegraphics[width=1\linewidth]{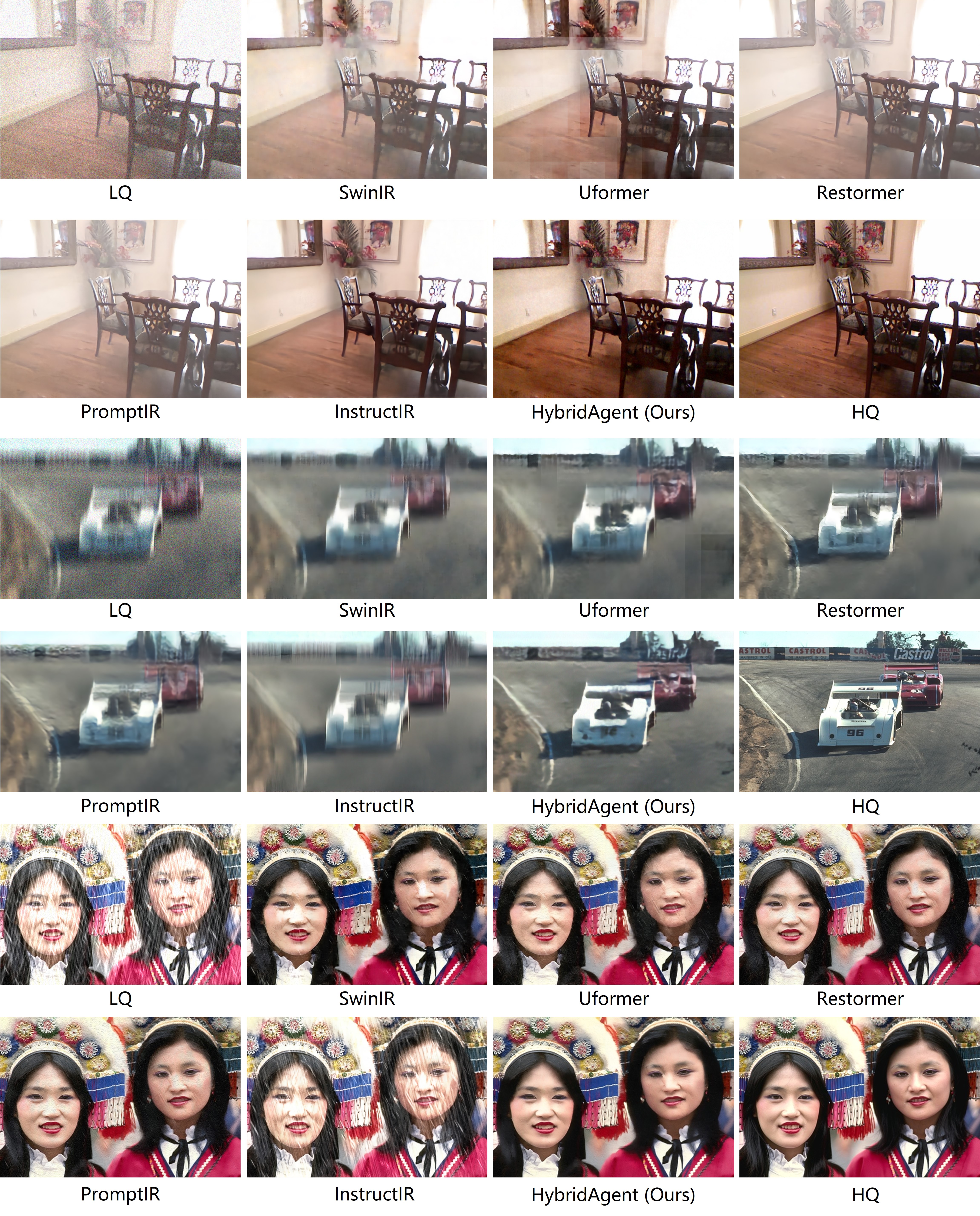}
    \caption{Qualitative comparisons between HybridAgent and other all-in-one methods. From top to bottom: Haze + Noise, Motionblur + Noise, Rainstreak + Noise.}
    \label{fig:aball-in-one}
\end{figure*}

In this section, we provide the qualitative comparisons between HybridAgent and other all-in-one methods to support the results in Table 4 of the main paper. As demonstrated in Figure~\ref{fig:aball-in-one}, HybridAgent achieves better restoration qualities than other methods, demonstrate the effectiveness of the collaboration of hybrid restoration tool and task-specific restoration tools.

\end{document}